\pdfoutput=1
\PassOptionsToPackage{table, x11names, dvipsnames}{xcolor}

\documentclass[11pt]{article}

\usepackage[]{acl}

\usepackage{times}
\usepackage{latexsym}

\usepackage[T1]{fontenc}

\usepackage[utf8]{inputenc}

\usepackage{microtype}

\usepackage{inconsolata}

\usepackage[toc,page,header]{appendix}
\usepackage{minitoc}

\usepackage{enumitem}

\newcommand{\amt}[1]{AMT-{#1}}
\newcommand{\blog}[1]{BLOG-{#1}}

\newcommand{\myparagraph}[1]{\par\noindent\textbf{{#1}.}} %

\newcommand \argmax {\operatorname*{arg\,max}} %

\newcommand{\method}{\textsc{JamDec}\xspace}
\newcommand{\ourmethod}{\textsc{JamDec}\xspace}

\newcommand{\cdbs}{\text{CoDi-BS}\xspace}

\newcommand\ximing[1]{{\color{brown}[\textit{#1}]$_{-XL}$}}

\usepackage[T1]{fontenc}
\usepackage[utf8]{inputenc}
\usepackage{latexsym}
\usepackage{amsmath}
\usepackage{amssymb}
\usepackage{bm}
\usepackage{bbm}
\usepackage{graphicx}
\usepackage{breakcites}
\usepackage{amsthm}
\usepackage{mdframed}
\usepackage{lipsum}
\usepackage{algorithm}  
\usepackage[noend]{algpseudocode}
\usepackage[english]{babel}
\usepackage{delimset} %
\usepackage{tikz}

\usepackage{url}            %
\usepackage{booktabs}       %
\usepackage{amsfonts}       %
\usepackage{microtype}      %
\usepackage{wrapfig}
\usepackage{caption}

\usepackage{minitoc} %
\usepackage{xspace}
\usepackage{enumitem}
\usepackage{cleveref}

\usepackage{amsthm}
\usepackage{graphicx}
\usepackage{latexsym}
\usepackage{mathtools}
\usepackage{multirow}
\usepackage{fancyvrb}
\usepackage{adjustbox}

\crefname{section}{Section}{Sections}
\crefname{appendix}{Appendix}{Appendices}

\crefname{theorem}{Theorem}{Theorems}
\crefname{lemma}{Lemma}{Lemmas}
\crefname{corollary}{Corollary}{Corollaries}
\crefname{proposition}{Proposition}{Propositions}
\crefname{definition}{Definition}{Definitions}
\crefname{assumption}{Assumption}{Assumptions}

\Crefname{algorithm}{Algorithm}{Algorithms}
\crefname{figure}{Figure}{Figures}
\crefname{table}{Table}{Tables}

\usepackage{url}            %
\usepackage{booktabs}       %
\usepackage{amsfonts}       %
\usepackage{nicefrac}       %
\usepackage{microtype}      %
\usepackage{xcolor}         %

\usepackage{amsmath}
\usepackage{amssymb}
\usepackage{xspace}
\usepackage{enumitem}
\usepackage{bbm}
\usepackage{lipsum}  
\usepackage{multirow}
\usepackage{booktabs}
\usepackage{eqparbox}
\usepackage{scalerel}
\usepackage{float}
\usepackage{wrapfig}
\usepackage{caption}
\usepackage{comment}
\usepackage{subcaption}
\usepackage{tabularx}
\usepackage{algorithm}
\usepackage{algpseudocode}
\usepackage{ragged2e}

\definecolor{platinum}{rgb}{0.9, 0.89, 0.89}

\usepackage{amssymb}%
\usepackage{pifont}%
\newcommand{\cmark}{\ding{51}}%
\newcommand{\xmark}{\ding{55}}
\usepackage{pdfpages}

\title{
\ourmethod: Unsupervised Authorship Obfuscation \\
using Constrained Decoding over Small Language Models
}

\author{Jillian Fisher$^{1}$\hspace{.3cm}Ximing Lu$^{1,2}$\hspace{.3cm}Jaehun Jung$^{1}$\hspace{.3cm}Liwei Jiang$^{1,2}$\hspace{.3cm}
\\
\textbf{
    Zaid Harchaoui$^{1}$\hspace{.3cm}Yejin Choi$^{1,2}$
} 
\\[0.1cm]
\hspace{0.8cm}$^{1}$University of Washington \hspace{0.3cm}$^{2}$Allen Institute for Artificial Intelligence \\[0.1cm]
\texttt{ jrfish@uw.edu} \\ 
}

\begin{document}
\maketitle

\begin{abstract}

The permanence of online content combined with the enhanced authorship identification techniques calls for stronger computational methods to protect the identity and  privacy of online authorship when needed, e.g., blind reviews for scientific papers, anonymous online reviews, or anonymous interactions in the mental health forums.  
In this paper, we propose an unsupervised inference-time approach to authorship obfuscation to address the unique challenges of authorship obfuscation: lack of supervision data for diverse authorship and domains, and the need for a sufficient level of revision beyond simple paraphrasing to obfuscate the authorship, all the while preserving the original content and fluency. 

We introduce \ourmethod, a  user-controlled, inference-time algorithm for authorship obfuscation that can be in principle applied to any text and authorship. Our approach builds on small language models such as GPT2-XL in order to help avoid disclosing the original content to proprietary LLM's APIs, while also reducing the performance gap between small and large language models via algorithmic enhancement. The key idea behind our approach is to boost the creative power of smaller language models through constrained decoding, while also allowing for user-specified controls and flexibility.
Experimental results demonstrate that 
our approach based on GPT2-XL outperforms previous state-of-the-art methods based on comparably small models, while performing competitively against GPT3.5 175B, a propriety model that is two orders of magnitudes larger.

 \end{abstract}

\section{Introduction}

 \begin{figure}[ht]
\centering
\includegraphics[scale=.3]{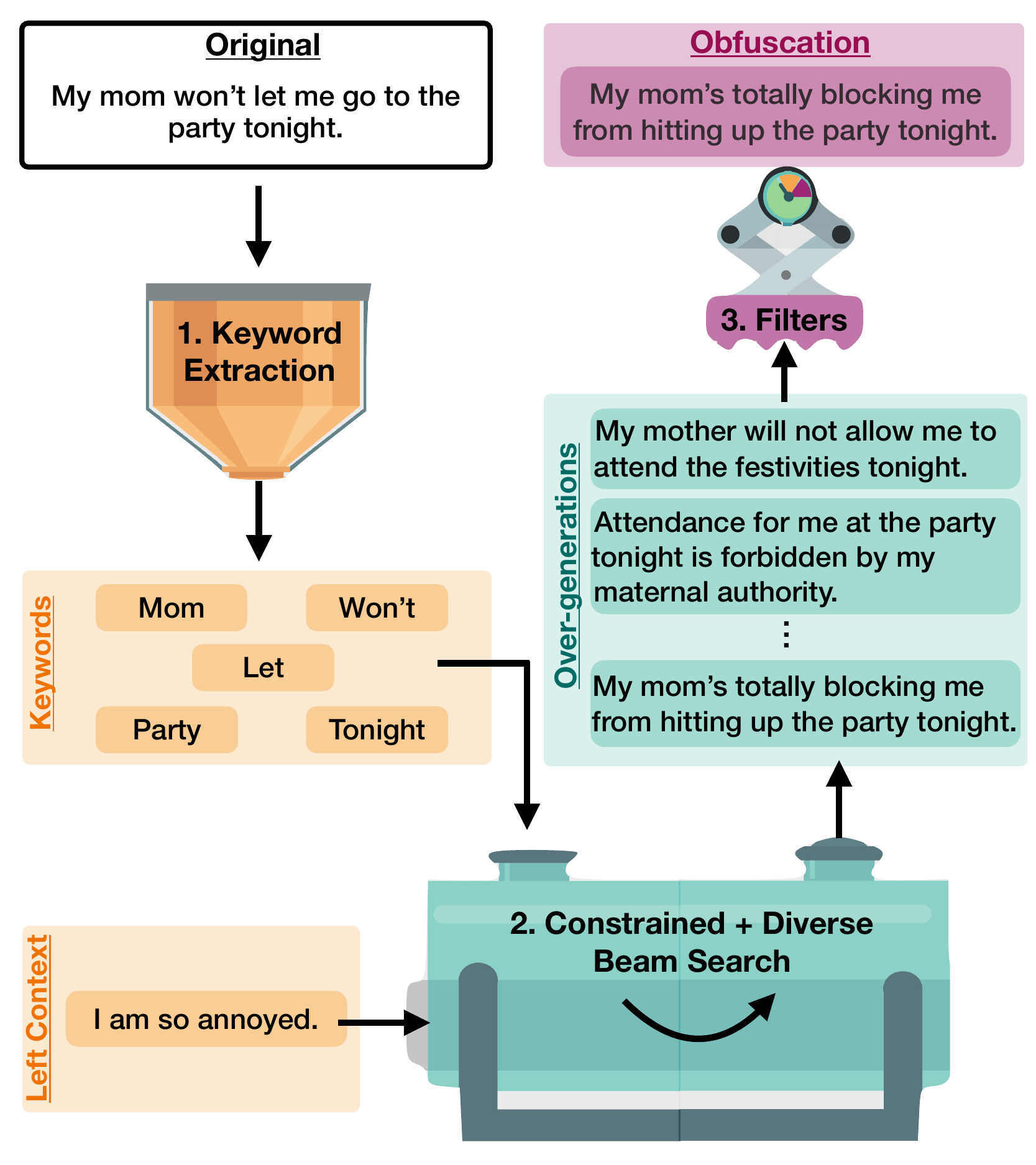}
\caption{\textbf{\ourmethod framework.} }
\label{fig:constrained-decoding-pipeline}
\end{figure}

Authorship obfuscation, the task of rewriting a text to protect the original writer's identity, has become increasingly important given the permanence of online content combined with new enhanced authorship attribution techniques \cite{ bright2021should, altakrori-etal-2022-multifaceted}. This task holds implications in various domains, including online privacy, and blind review in academic research.
However, safeguarding an authorship style, while maintaining the same content and grammatical fluency, is a complex task.

Unlike other authorship-related tasks such as paraphrasing or style transfer, authorship obfuscation poses unique technical challenges due to its different assumptions. For example, paraphrasing involves rephrasing an original text, but can be accomplished without altering the original style. Conversely, for style transfer, the task requires a predetermined target style. However, in the case of authorship obfuscation, there is no fixed endpoint style to guide the generation because the main goal is the absence or avoidance of a particular style. In fact, it may involve incorporating multiple styles or navigating a wide spectrum of possibilities to achieve success.\footnote{A more detailed discussion on the differences between these authorship-tasks can be found in \cref{appx:compare_author_tasks}.}

One approach to authorship obfuscation is to use large language models, such as ChatGPT or GPT4. However, these models require large computing resources. Furthermore, if a user employs a method based on proprietary LLMs that retain user data, they are vulnerable to extra privacy threats or the leakage of their original content. To mitigate these risks, non-model or smaller closed model methods are preferred.

Other previous approaches for authorship obfuscation include the use of round-trip machine translation \cite{Keswani2016AuthorMT}, strict rule-based algorithms \cite{stylo_method}, or iterative-change algorithms \cite{mahmood2019mutantx}. However, these methods either do not lead to enough modification \cite{Keswani2016AuthorMT}, diverge into grammatically incorrect text due to the rigid rules \cite{stylo_method}, or require an additional large-scale authorship corpus \cite{mahmood2019mutantx}. Therefore, in comparison to modern LLMs, we find a notable performance gap between previous methods developed for smaller models. 

To overcome these limitations, we present \ourmethod, a light-weight, user-controlled, unsupervised inference time algorithm for authorship obfuscation that can be used with any arbitrary text. \ourmethod employs smaller base models such as GPT2, which by themselves are too weak to produce accurate paraphrases, let alone obfuscation \cite{Jung2023ImpossibleDF}. To overcome this weakness, we frame the task as a constraint decoding problem, where the constraint is given as lexical keywords to include to control the content of the generation. To identify these keywords automatically, we leverage likelihood scores from smaller models. Lastly, since the decoded text is not guaranteed to be faithful to the original text, we design a filtering step that can be uniquely adjusted by the user. An overview of \ourmethod three-stage framework can be found in  \cref{fig:constrained-decoding-pipeline}. The name is inspired by Jambalaya, the popular American Creole and Cajun rice dish which is a mixture of meat, vegetables and spices. 

We provide experimentation on two datasets, scholarly articles and diary-style entries with a range of three to ten authors. The results show that \ourmethod performs better than state-of-the-art methods of similar size and comparable to significantly larger language models in both automatic and human evaluations. In particular, we demonstrate that \ourmethod is able to obfuscate, while simultaneously preserving the original content, which previous methods cannot achieve. \footnote{Code is available:\href{https://github.com/jfisher52/JAMDecoding}{https://github.com/jfisher52/JAMDecoding}}

\begin{table*}[t!]\centering
    \resizebox{\textwidth}{!}{
    \begin{tabular}{ llcccccccc }\toprule
        & \multicolumn{1}{c}{\textbf{Method}} & \multicolumn{2}{c}{Mutant-X} & \multicolumn{1}{c}{Paraphrase}& \multicolumn{1}{c}{Machine Transl.}& \multicolumn{1}{c}{Stylometric}& \multicolumn{2}{c}{\ourmethod}\\
        
        \cmidrule(lr){3-4}\cmidrule(lr){5-5}\cmidrule(lr){6-6}\cmidrule(lr){7-7}\cmidrule(lr){8-9}
        
        \multicolumn{1}{c}{\textbf{Dataset}}&\multicolumn{1}{c}{\textbf{Metric}} & \textit{ENS} & \textit{RFC}  & & & & \textit{W/O Stylo} & \textit{W/ Stylo}  \\
        
        \midrule
        &Drop Rate (ENS) & $\star$ &	-0.04 &	 \underline{0.04} &	\underline{0.04} &	-0.03 & \textbf{0.11} & \textbf{0.11}	\\
        &Drop Rate (BertAA) & \underline{0.10} &	0.04 & 0.04 &0.08 &	\textbf{0.12} & 0.04 & 0.04	\\
        &METEOR& \underline{0.80}&	\textbf{0.81}& 0.55	&0.69&	 \underline{0.80} & 0.62 & 0.62\\
      \amt{3}  &NLI&       0.60&	0.61&	 0.62	&\underline{0.75}&	0.50& \underline{0.75} &  \textbf{0.81}\\
      &CoLA&       0.50&	0.51 &	0.78&0.69&	0.46 &\textbf{ 0.85} & \underline{0.79}\\
         \rowcolor{platinum}   &Task Score (ENS)&  $\star$   &	0.36	& 0.48&\underline{0.49}	 &0.31 & \textbf{0.57} & \textbf{0.57} \\
                    \rowcolor{platinum}        &Task Score (BertAA)&    0.40  & 	0.39	&0.48 &	\underline{0.51} & 0.36 & \textbf{0.55} &  \textbf{0.55}\\
        \midrule
        &Drop Rate (ENS) &	$\star$ &	0.08& \underline{0.20}&	\underline{0.20}&	\textbf{0.23} & 0.10 & 0.13\\
        &Drop Rate (BertAA) & \underline{0.07} &	0.00  &-0.06 &	\underline{0.07} &	0.04 & \textbf{0.14} & \textbf{0.14}	\\

                &METEOR &    \underline{0.74}&	0.72& 0.57 &	0.68 & \textbf{0.79} &	0.61 & 0.61\\
 \amt{5}&NLI  &      0.56&	0.57 & 0.62&  0.74 & 	0.48& \underline{0.76} &  \bf{0.82}	\\
       &CoLA   &       0.51&	0.55 & 0.77& 0.69& 	0.46& \textbf{0.85} &  \underline{0.79}\\
                \rowcolor{platinum}   &Task Score (ENS)&  $\star$ &	0.40 & 0.53&0.54	&0.39 & \textbf{0.57} & \underline{0.58}\\
                \rowcolor{platinum}        &Task Score (BertAA)&  0.38 &0.37  & 0.44&	\underline{0.50} &	0.33 & \textbf{0.58} & \textbf{0.58}	\\
        \midrule
        &Drop Rate (ENS) &	$\star$ &	0.10 & 0.07&0.19	&0.11 & \textbf{0.44} & \underline{0.41}	\\
         &Drop Rate (BertAA) & 0.03 &	\underline{0.04}  & -0.04&	\textbf{0.06} &	0.00 & -0.03 & -0.02	\\
                 &METEOR &	\underline{0.84}& \textbf{0.86}& 0.54& 0.66 & 0.81	& 0.60 & 0.61		\\
       \amt{10}&NLI	&  0.61    &0.64& 0.61&\underline{0.73} & 0.45	& \textbf{0.79} & \textbf{0.79}\\
        &CoLA	&    0.53 & 0.57& \underline{0.77}&0.68 & 0.46  & \textbf{0.78} & \textbf{0.78}\\
                 \rowcolor{platinum}   &Task Score (ENS)& $\star$ & 0.44	 & 0.48 & 0.53	&	0.34 &\textbf{0.67} & \underline{0.66} \\
                          \rowcolor{platinum}        &Task Score (BertAA)&   0.39	 & 0.42 & 0.45	&	0.49 &0.30 & \underline{0.51} & \textbf{0.52} \\
        \midrule
        \midrule
        &Drop Rate (ENS) & $\star$& \underline{0.28}	&\textbf{0.31} &0.18	&0.03 & 0.03	&0.03	\\
         &Drop Rate (BertAA) & 0.06 &	\underline{0.30} & \textbf{0.47}&	0.0 &	0.0 & 0.29 & 0.29	\\
                &METEOR &	\underline{0.79}& 0.59& 0.44& 0.58 & \textbf{0.82}	& 0.53	&0.52	\\
       \blog{5}  &NLI	&  0.58    &0.47&0 .49&0.65 & \bf{0.75}	&	\underline{0.68}&\underline{0.68}	\\
        &CoLA	&    0.44 & 0.46& 0.63&0.55 & 0.44  &	\textbf{0.74}&\underline{0.73}	\\
                 \rowcolor{platinum}   &Task Score (ENS)&    $\star$   &	0.40 & \underline{0.47} &	0.46&	0.41 & \bf{0.48} &\textbf{0.48} \\
                          \rowcolor{platinum}        &Task Score (BertAA)&  0.36   & 0.41	&\underline{0.53} &	0.40& 0.40&\textbf{0.57} & \textbf{0.57} \\
        \midrule
        \midrule
        &Drop Rate (ENS) &	$\star$& 0.13 & \textbf{0.35 }&0.30	&0.21& 0.23&\underline{0.32}\\
         &Drop Rate (BertAA) & \underline{0.37} &	0.06 & \textbf{0.40}&	0.11 &	0.08 & 0.32 & 0.32	\\
                 &METEOR &	0.55& \textbf{0.85}&0.43 & 0.61 & \underline{0.82}	&0.54	&0.53	\\
      \blog{10} &NLI	&  0.46    &0.61& 0.46& 0.62 & \textbf{0.75}	&	\underline{0.67}& \underline{0.67}	\\
        &CoLA	&    0.47 & 0.45&\underline{0.62} &0.54 & 0.41  &	\textbf{0.74}&\bf{0.74}	\\
                 \rowcolor{platinum}   &Task Score (ENS)&   $\star$   &0.40	 & 0.48 &0.49	&0.46	 & \underline{0.55} &\textbf{0.58} \\
                          \rowcolor{platinum}        &Task Score (BertAA)&  0.43   & 0.37	& 0.49 & \underline{0.42} & 0.41& \textbf{0.58} & \textbf{0.58}\\
        \midrule
        \bottomrule
    \end{tabular}
    }
    \caption{Results from the automatic evaluation for Mutant-X (using two internal classifiers; ENS and RFC), GPT3, Paraphrasing, Machine Translation, Stylometric and \ourmethod (using two variation of filtering; with and without stylometric-based obfuscator Stylo) across all datasets. The \textbf{highest} value is bolded and the \underline{second-highest} value is underlined. Methods that use the same evaluation classifier during obfuscation are excluded ($\star$).}
    \label{tab:amt_class_metric}
\end{table*}

\section{Background on Authorship Obfuscation}
\myparagraph{Setup}
Let $\mathcal{A}$ be a given set of authors. We consider an original text $y_{\text{orig}}$ that was written by author $B \in \mathcal{A}$. The task of authorship obfuscation aims to create a new text $y_{\text{obf}}$ which can not be identified as written by author $B$. For evaluation, we consider a classification model $M(\cdot)$ (also known as an authorship attribution models), which has been trained to classify texts of each author in $\mathcal{A}$. The aim is to create a method $f(\cdot)$ such that $M(f(y_{\text{orig}})) \neq B$.

\myparagraph{Measure of a Successful Algorithm}
Our goal is to create an obfuscated version of the original text that preserves the meaning and intent of the original text, while making it difficult to attribute the authorship to the original author. Following past literature \cite{mahmood2019mutantx, PAN2018, altakrori-etal-2022-multifaceted}, we consider an obfuscation method successful if the obfuscated text satisfies the following three requirements:
\setlist{nolistsep}
\begin{itemize}[noitemsep, align=left, labelindent=.05em, leftmargin=*]
\itemsep0em 
  \item \textbf{Style Concealment} Analysis of the obfuscated text does not reveal the original author. This is usually measured using an authorship attribution model or a threat model \cite{Mahmood2019AGH}.
  \item \textbf{Content Preservation} The content of the original text is maintained. Metrics such as METEOR \cite{lavie-etal-2004-significance}, and Natural Language Inference models (NLI) \cite{liu-etal-2022-wanli} can be used to measure content overlap.
  \item \textbf{Language Quality} The obfuscated text is grammatically correct and natural sounding. Grammaticality of a text can be measured using a Corpus of Linguistic Acceptability (CoLA) model \cite{cola}. Text fluency can be determined using human evaluation. 
\end{itemize}

\myparagraph{Inference-time Algorithms for Authorship Obfuscation}
To address this task, we propose using an inference time algorithm that can obfuscate a text on-the-fly, rather than training a model on a specific author's writing style. We choose to use a decoding time algorithm over fine-tuning as it offers several benefits, including more flexibility in the generation and the ability to obfuscate text without access to a corpus of the author's writing.

Our proposed algorithm draws inspiration from various sources, including Diverse Beam Search \cite{vijayakumar2018diverse}, Lexically Constrained Decoding \cite{post2018fast}, and Neurologic decoding \cite{neurologic}.

\section{\ourmethod}\label{sect:method}

 We present \method, which obfuscates any text without any prior knowledge of the author. \method is composed of three main steps: keyword extraction, over-generation, and filtering, which can be implemented on a sentence, paragraph, or full document level. 

\subsection{Step 1: Keyword Extraction}
First, we identify crucial keywords that encapsulate the original text's content, and later ensure its inclusion in the generated obfuscated text to maintain content preservation. We explore multiple keyword extraction methods, including embedding-based extraction and likelihood-based extraction.

\myparagraph{Embedding-based method} KeyBERT is a popular method for keyword extraction \cite{grootendorst2020keybert}, which uses BERT-embeddings and cosine similarity to find the sub-phrases in a document that are the most similar to the document itself.

\myparagraph{Likelihood-based method} At a high level, we select the top-$k$ tokens with the lowest conditional probabilities, as measured by a specific language model, as keywords for a given sentence. Intuitively, these tokens represent content that a language model might most struggle to generate accurately. We experiment with both an auto-regressive language model GPT2, and text-to-text language model T5. For GPT2, we compute the likelihood of each token conditioned on its previous content. For T5, we leverage its fill-in-the-blank ability by providing an input sentence with a specific token masked. We then calculate the probability of T5 generating that particular token as the infill, which serves as the likelihood of that token.

Since all the methods yield valid keywords in practice (see \cref{appx:keyword_extraction_comparison}), we utilize them all to generate numerous candidates for subsequent filtering to achieve high-quality obfuscation.

\subsection{Step 2: Over-Generating Candidate Obfuscations}
Next, we utilize the previously extracted keywords and the left context of $y_{\text{orig}}$ to over-generate many variations of $y_{\text{orig}}$. We use $m$ sentences occurring before $y_{\text{orig}}$ as the left context to encourage fluid generation. Our goal is to produce multiple generations constrained by the extracted keywords, ensuring content similar to $y_{\text{orig}}$. At the same time, we aim to produce a variety of generations with diverse authorship styles to achieve obfuscation effectively.
To achieve these seemingly opposing goals, we merge two decoding techniques, Lexically Constrained Beam Search \cite{post2018fast}and Diverse Beam Search \cite{vijayakumar2018diverse}, and refer to the combined approach as Constrained Diverse Beam Search (\cdbs). 

\myparagraph{Constrained Diverse Beam Search}
\cdbs employs Constrained Beam Search (Co-BS) as the base algorithm, but uses the scoring function from Diverse Beam Search (Di-BS) instead of likelihoods when iteratively selecting the top $k$ candidates from each bank. Its objective function can be represented as: 
\vspace{-1.5mm}
\begin{align*}
    \argmax_{w \in W} P_w(y\vert x) + \lambda_1 D(y,Y) + \lambda_2 C(y)
\end{align*}
where $x$ is the sequence of previous tokens, $D(y, Y)$ is a diversity term measuring the dissimilarity between the output sequence $y$ and the set of previously selected sequences $Y$ within the beam, $C(y)$ is a constraint function quantifying the degree to which the output sequence $y$ satisfies the constraints, $\lambda_1, \lambda_2$ are hyperparameters controlling the weight of the diversity and constraint penalty, and $w \in W$ is the parameter vector. Intuitively, \cdbs promotes candidates distinct from the previously chosen ones, while also ensuring that they satisfy a specific number of constraints. \cref{appx:algorithms} has an overview of the \cdbs algorithm and details of both Constraint and Diverse Beam Search separately.

\subsection{Step 3: Filtering Candidate Obfuscations}\label{sect:filtering_stage}
The filtering stage comprises multiple steps to refine the pool of candidates from the previous stage, ultimately choosing the most suitable obfuscation. This step enables the user to have full control in selecting generations based on any metric. In our pipeline, we first filter based on an NLI (Natural Language Inference) threshold, which evaluates the coherence and content overlap between the generations and the original text. Next, we further filter the remaining candidates based on a CoLA (Corpus of Linguistic Acceptability) threshold, which focuses on the grammatical correctness and linguistic acceptability of the generations. Finally, and optionally, taking into account any previous knowledge of the author, we choose the ultimate obfuscation to be the generation that deviates the most from the original author's style. In our experiment, we do not assume any prior knowledge of the authors to showcase the effectiveness of our method in a more challenging situation.

 %
\section{EXPERIMENTS}
\label{sec:expt}
We evaluate two versions of \method on two benchmarks in distinct domains: scholarly passages and diary-style entries. For baselines, we consider three state-of-the-art methods for authorship obfuscation: Mutant-X \cite{mahmood2019mutantx}, Round-Trip Translation \cite{Keswani2016AuthorMT}, and Stylometric \cite{stylo_method}, and a paraphrasing method \cite{pegasus}. As a stronger baseline, we also consider using zero-shot prompting of GPT3.5 175B which is orders of magnitude larger \cite{gpt3}. For further details, see \Cref{appx:exp_details} and for access to the code see here.

\begin{figure}
\begin{flushleft}
\includegraphics[width=0.5\textwidth]{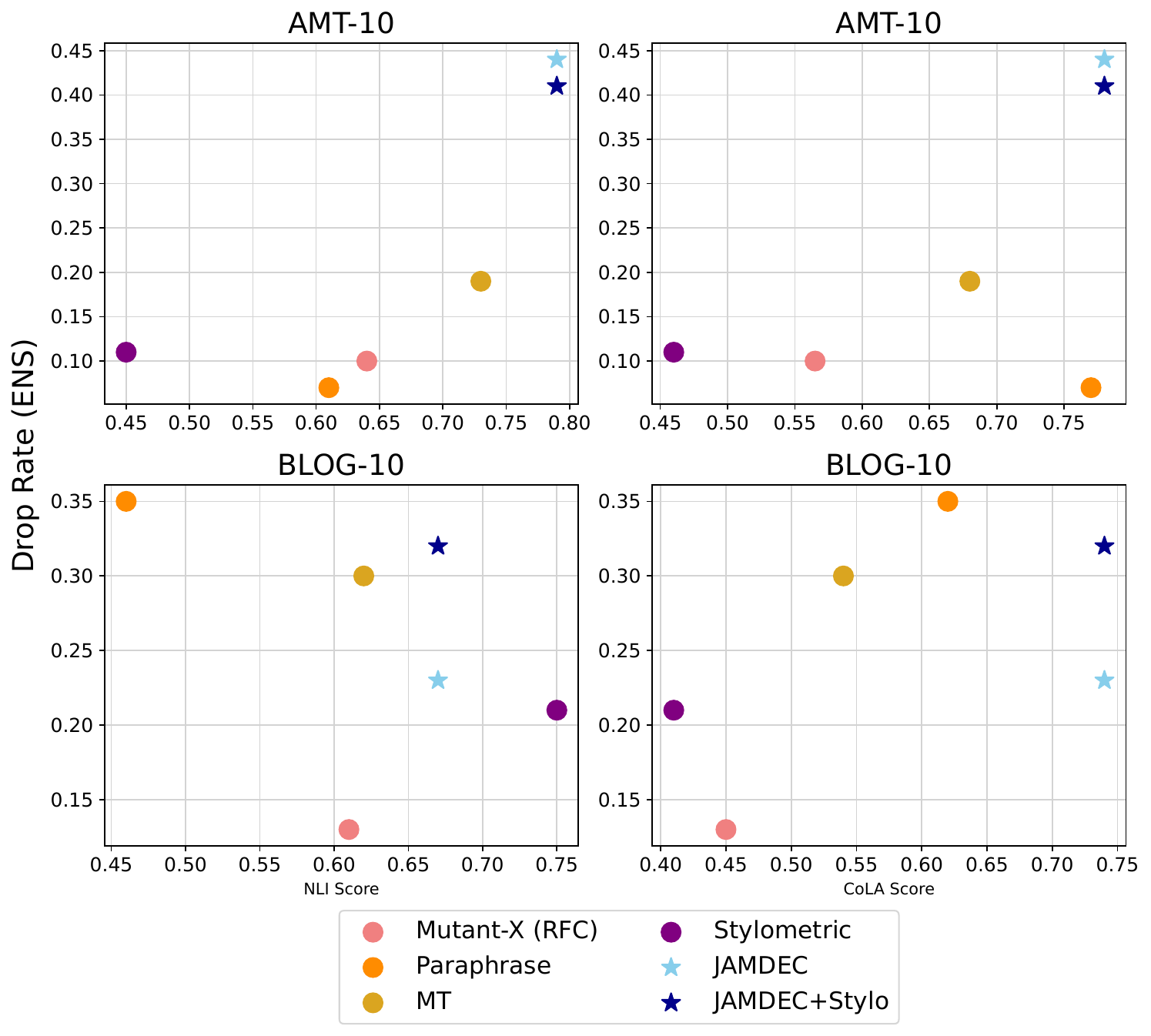}
\caption{Highlighting the trade-offs between obfuscation (Drop Rate (ENS)), content preservation (NLI), and language quality (CoLA) of each method for the AMT-10 and BLOG-10 datasets. The dotted line indicates the trend through all methods.}
\label{fig:nli_cola_tradeoff}
\end{flushleft}
\end{figure}

\subsection{Setup}
\myparagraph{Datasets}
We used two datasets to evaluate \method. The first is the Extended-Brennan-Greenstadt \cite{amt_dataset} which is a collection of "scholarly" short (~500-word) paragraphs gathered from Amazon Mechanical Turk (AMT).
We use this dataset, which we refer to as AMT, to produce three test datasets with 3, 5, and 10 authors, with $n= 27, 30, 49$ texts respectively (\amt{3}, \amt{5}, \amt{10}). 

The second dataset is the Blog Authorship corpus \cite{blog_dataset}, a collection of blogs (diary-style entries) that were posted to blog.com. 
Similarly, we use this dataset to construct two datasets with 5 and 10 authors, with $n = 72, 150$ texts respectively (\blog{5}, \blog{10}). 

\myparagraph{\method Configuration} \xspace
To promote diversity of generated candidates, we employ all three types of keyword extraction methods, (KeyBERT, Likelihood-GPT2, and Likelihood-T5), and either \cdbs or only CBS. We ran with a beam width of 50. All other details can be found in \cref{appx:exp_details}. 

In the filtering stage, we occasionally find cases where none of the generations passes either NLI or CoLA filter. We consider two ways of handling such cases -- (1) \ourmethod, where we simply output the original sentence, (2) \textsc{\ourmethod + Stylo}, where we run a basic stylometric obfuscator on the original sentence.\footnote{The detail of the basic stylometric obfuscator is provided in \cref{appx:stylometric-based-obfuscator}.}
\definecolor{ForestGreen}{RGB}{34,139,34}
\definecolor{Maroon}{RGB}{100, 0,0}
\begin{figure}[h!]
\begin{flushleft}
\includegraphics[width=.5\textwidth]{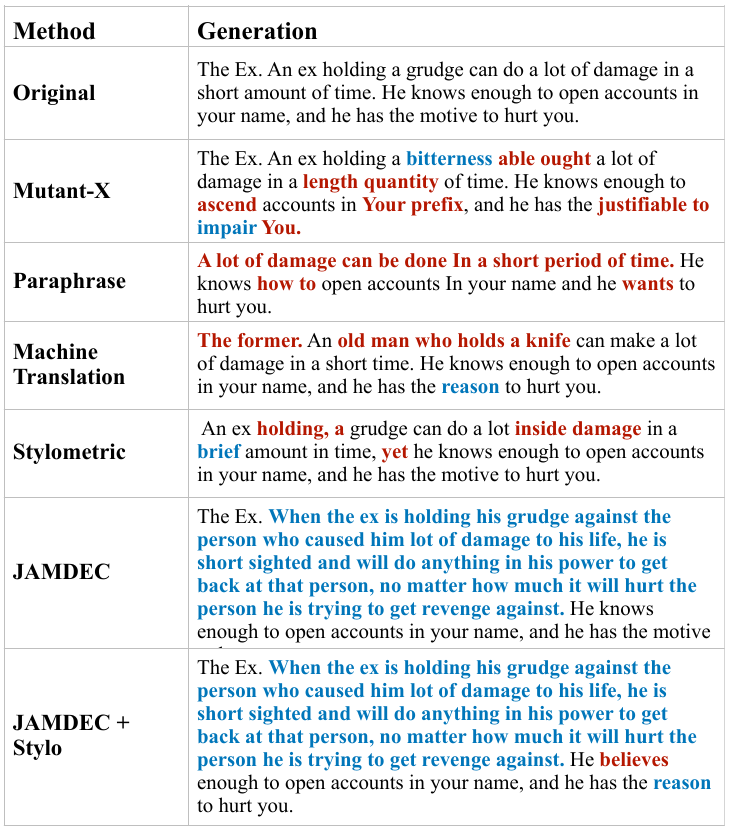}
\caption{Qualitative examples of obfuscated text created by each method. The sentences are taken from the AMT-3 dataset. Changes to the original are outline in \textcolor{teal}{blue} (correct grammatically and in context) and \textcolor{Maroon}{red} (incorrect grammatically or out of context).}
\label{fig:qual_examples}
\end{flushleft}
\end{figure} 
\myparagraph{Baselines}\footnote{An additional baseline, Style Transfer, can be reviewed in \cref{appx:style_transfer}.} \xspace 
We use the following baselines.

\textit{Stylometric Obfuscation: } 
A stylometric obfuscation (Stylometric) proposed by \citet{stylo_method}, calculates a suite of statistical features (e.g. average number of words per sentence, word frequency, etc.) that are indicative of style, then modifies the text such that these metrics align with an "average" value, pre-calculated on a training set.

\textit{Mutant-X: } Mutant-X \cite{mahmood2019mutantx} is a genetic algorithm which iteratively substitutes words in the original text with the synonyms selected by an internal classifier. Additionally, at random iterations, it incorporates a "crossover" effect that involves cutting two parent texts at a random position and combining them to create two new child texts. This method does require an additional authorship corpus to train the internal classifier. For consistency, we adopt the same features and architectures for the internal classifier (Ensemble and Random Forest), as suggested in the subsequent work by Haroon et al. \cite{avengers}. For more information on training these classifier models, reference \cref{sect:evaluation_details}. To accurately compare with all methods, we leave out any results from Mutant-X where the internal classifier matches the evaluation classifier, since we do not assume access to the evaluation models during obfuscation. 

\textit{Paraphrasing: } Although paraphrasing has a slightly different goal than authorship obfuscation, we include the comparison for a thorough investigation of all methods. We employ a state-of-the-art paraphrasing model, PEGASUS Paraphrase \cite{pegasus, paraphrase_model} a PEGASUS model fine-tuned on a self-supervised task for paraphrasing. 

\textit{Round-Trip MT: } Additionally, we consider a baseline powered by round-trip translation, a popular approach for authorship obfuscation \cite{Keswani2016AuthorMT}. We implement the approach using M2M100, a state-of-the-art translation model, translation English text into German, then to French, and finally back to English.

\textit{GPT3.5: } Lastly, considering the significant progress made in large language models, we include a comparison with zero-shot prompted GPT3.5 (text-davinci-003) \cite{gpt3}. We consider two approaches -- sentence-level obfuscation (obfuscating each sentence individually), and paragraph-level obfuscation (obfuscating the entire text as a whole). We note that prompt selection is very important and tried to find the best prompt for the task. The specific prompts utilized for this purpose can be found in \cref{appx:exp_details}. Due to financial constraints, we limit this baseline to AMT-3.

A time consumption analysis of these methods can be found in \cref{appx:time_consumption}.
\begin{table}[h]\centering
    \resizebox{.5\textwidth}{!}{
    \begin{tabular}{ lcccc }\toprule
        \multicolumn{1}{c}{\textbf{Method}}  & \multicolumn{2}{c}{GPT3.5} & \multicolumn{2}{c}{\ourmethod}\\
        
        \cmidrule(lr){2-3}\cmidrule(lr){4-5}
        
        \multicolumn{1}{c}{\textbf{Metric}} & \textit{Sentence} & \textit{Paragraph} & \textit{W/O Stylo} & \textit{W/ Stylo}  \\
        
        \midrule
        Drop Rate (ENS) & \textbf{0.23} &	\textbf{0.23} & \underline{0.11} & \underline{0.11}	\\
        Drop Rate (BertAA) & \textbf{0.13} &	\underline{0.09} & 0.04 & 0.04	\\
        METEOR& 0.33&	\underline{0.41}& \textbf{0.62} & \textbf{0.62}\\
      NLI& \underline{0.77}&	0.73 & 0.75 &  \textbf{0.81}\\
      CoLA&0.76&	\underline{0.80} &	\textbf{ 0.85} & 0.79\\
         \rowcolor{platinum}   Task Score (ENS)&  \textbf{0.59}	& \textbf{0.59}& \underline{0.57} & \underline{0.57} \\
                    \rowcolor{platinum}        Task Score (BertAA)&	\textbf{0.55} & \underline{0.54}	& \textbf{0.55} &  \textbf{0.55}\\
        \bottomrule
    \end{tabular}}
    \caption{Results from the automatic evaluation for GPT3 and \ourmethod (using two variation of filtering; with and without stylometric-based obfuscator Stylo) for AMT-3. The \textbf{highest} value is bolded and the \underline{second-highest} value is underlined. }
    \label{tab:gpt3_all_metrics}
\end{table}

\myparagraph{Automatic Evaluation}\label{sect:evaluation_details}
We evaluate all method along the following three axes.

    \textit{\textit{1. Style Concealment: }}  In line with past work, we use two authorship attribution models trained on stylometric features for authorship verification. The first employs Writeprints-static \cite{amt_dataset}, a collection of lexical and syntactic features, such as word length, average word count, and usage of function words, among others. Recognizing that classification from one model may not transfer effectively to all text \cite{mahmood2019mutantx}, we adopt the ensemble attribution classifier (ENS) methodology introduced by \citet{avengers}, which comprises several attribute-based classifiers, each utilizing different attributes, and leverages a voting system for their aggregation. Since this has been shown to give the most accurate classification results \cite{avengers}, we use ENS for both the Mutant-X method and evaluation. We also train a random forest classifier (RFC) as another internal classifier for the Mutant-X method. Further details on the training can be found in \cref{appx:exp_details}.

    Second, we use a more sophisticated architecture by using BertAA model \cite{fabien-etal-2020-bertaa}, a BERT fine-tuned specifically for authorship attribution. \footnote{A comparison of these authorship attribution models can be found in \cref{tab:classifier_acc}.} 

    Using an authorship attribution model (either ENS or BertAA), we calculate the \textit{Drop Rate} or the average drop in the percentage of obfuscated text that were not identified as the true author compared to the original text. This metric accounts for any inaccuracy of the attribution models, see \cref{appx:exp_hyperparam} for more information.

    We note that adversarial threat model can be used for further evaluation and comparison \cite{Zhai2022AGH, mahmood-etal-2020-girl}, therefore provide an ablation study in \cref{appx:threat_model} using this type of evaluation. 
    
    \textit{2. Content Preservation: } 
    To maintain consistency with previous studies, we compute the METEOR \cite{meteor} score between the original and obfuscated text, which evaluates token overlap \cite{mahmood2019mutantx, ant}. However, we note that content semantics can be preserved without direct token overlap by the use of synonyms, therefore we also assess the probability of entailment between the original and obfuscated text using a natural language inference (NLI) model called WANLI \cite{liu-etal-2022-wanli}. We will rely on NLI as the main component of content overlap due to its flexibility in measuring content preservation and coherence. 
    
    \textit{3. Language Quality: } To measure language quality, we employ a TextAttack \cite{textattack}, which fine-tunes RoBERTa \cite{roberta} on the Corpus of Linguistic Acceptability (CoLA) \cite{cola}. The CoLA dataset consists of 10.6k sentences that have been linguistically annotated to assess their grammatical correctness.    

\textit{Overall Task Score: } While each of the dimensions above is crucial for the holistic evaluation of author obfuscation system, we also aim to provide an aggregate of the scores into a single task score. Therefore, we also define \textit{Task Score}, an unweighted average of the Drop Rate (using ENS or BertAA), NLI score, and CoLA score. We use the mean of the dimension, as the task of authorship obfuscation is deemed to be successful only if all three goals are satisfied. \footnote{We also provide each scores individually in case the reader prefers to weight a certain goal more heavily.}:
\vspace{-3mm}
\begin{align*}
    \text{Task Score} = \frac{\text{Drop Rate}+ \text{NLI}+ \text{CoLA}}{3}.
\end{align*}

\begin{figure}
\begin{flushleft}
\includegraphics[width=0.5\textwidth]{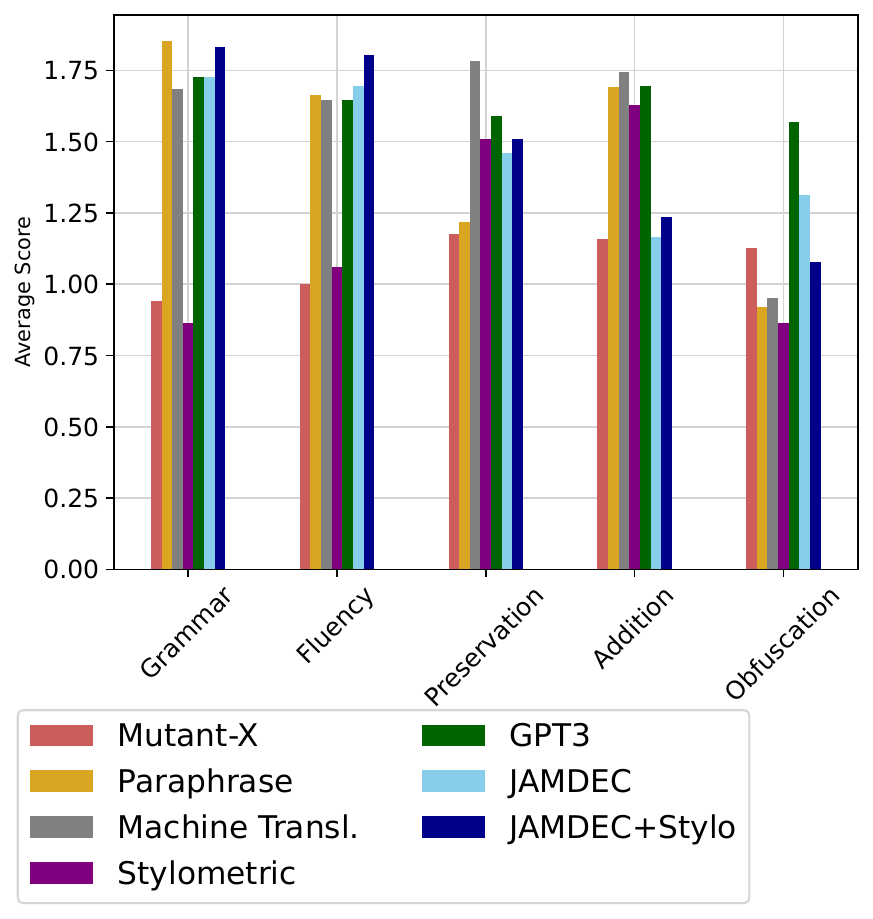}
\caption{Human Evaluation on 102 random samples from AMT-3. We include two versions of our method with differing filtering stages (with and without Stylo). }
\label{fig:human_eval}
\end{flushleft}
\end{figure}

\myparagraph{Human Evaluation}
On dataset AMT-3, we additionally use human evaluations to validate our automatic measures. We randomly select 102 short passages (one to four sentences) from AMT-3 for this evaluation. We employed Amazon Mechanical Turk workers to read both the original and obfuscated text, and then asked a series of five questions to be rated on a three-point likert scale.

\subsection{Main Results}
\myparagraph{\method has higher Task Score compared to all task-specific methods and similar or better to GPT3}
In \cref{tab:amt_class_metric} and \cref{tab:gpt3_all_metrics}, we present the results from the automatic evaluations. \ourmethod (with or without Stylo) with 1.5B GPT2-XL has the highest Task Scores for almost every dataset, and only $2\%$ lower BertAA Task Score than 175B GPT3.5. Of note, is \amt{10}, where it performs more than $10\%$ higher than almost all other methods on ENS and BertAA Task Score. This indicates, that \ourmethod is successful in all three goals of authorship obfuscation across different genre of texts. Also, we observe that the two variations of \ourmethod perform similarly across the datasets.   

\myparagraph{\ourmethod strikes a better balance between content preservation and author obfuscation} 
\cref{fig:nli_cola_tradeoff} depicts the variability in the \amt{10} and \blog{10} datasets' Drop Rate, NLI score, and CoLA score. Preferably, a method should score high in all metrics, resulting in a position in the top right quadrant of each graph. However, we observe a clear trade-off for each of the task-specific baselines. For example, in BLOG-10, the Paraphrase method has an ENS Drop Rate $3\%$ higher than  \ourmethod, but it also has a $12\%$ lower CoLA rate and $21\%$ lower NLI, as seen by the orange dots in the top left corner and center of the bottom left and right graph. In contrast, we observe that \ourmethod lies closely to the top right in each graph, demonstrating its effectiveness in balancing the various objectives of authorship obfuscation. Other datasets show similar results and can be viewed in \cref{appx:metric_comparison_all}. 

This is also supported by qualitative inspection, where we notice poor grammar quality in obfuscated text produced by the task-specific methods, which makes it easy to trick an automatic classifier, however does not maintain the quality and content of the original text. This was particularly relevant in the BLOG datasets, which already contains informal language that can be easily corrupted by single word replacement methods. We provide a qualitative example in \cref{fig:qual_examples}. 
\myparagraph{Human evaluations confirms that \ourmethod maintains language quality while successfully obfuscating} The outcomes of the human evaluation of AMT-3 are shown in \cref{fig:human_eval}. Similar to the automatic evaluation, \ourmethod human evaluation scores are $5\%-50\%$ higher for Grammar and Fluency, than most other method, including GPT3.5. For Content Preservation, \ourmethod performs on-par with GPT3.5, while Machine Translation unsurprisingly scores the highest because it only tends to slightly modify the original text, as shown in \cref{fig:qual_examples}. While we observe \ourmethod to be relatively weak in Content Addition, we attribute this mainly to the limitation of the human evaluation environment.
Our approach involves utilizing a left context in the beam search process, allowing the model to consider information from earlier sentences when generating subsequent ones.  As a result, some generations incorporate data from earlier sentences. However, the samples used for the human evaluation were random short passages taken from the whole text, making it possible for the workers to perceive the information as an "addition" when it was actually present earlier in the passage. However, despite this, we see that \ourmethod performs better than all task-specific methods in Obfuscation by at least $10\%$. 

\subsection{Ablation and Other Studies}
We conduct ablation studies \footnote{Full details in \cref{appx:additional_experiements}} on \ourmethod, to better understand the contribution of each component. 
\myparagraph{\ourmethod performs better at authorship obfuscation using \cdbs} We find that using \cdbs leads to an overall increase in Drop Rate of $\sim 6\%$ and an increase in the number of sentences that pass the base NLI and CoLA threshold of about $32\%$, with little change in NLI and CoLA score compared to only using CBS. 

\myparagraph{\textsc{\ourmethod + Stylo} performs better in human evals \textit{without} the CoLA threshold} We run an additional human evaluation with obfuscation created using \textsc{\ourmethod + Stylo} but \textit{without} a final CoLA threshold. Without a final CoLA threshold, all sentences transformed using Stylo were used. It resulted in an overall increase in Obfuscation of $0.09\%$ compared to \ourmethod+Stylo with a threshold, making it higher than all task-specific methods. However, it did have a decrease of $0.15\%$ and $0.13\%$ in Grammar and Fluency, respectively.

\myparagraph{\ourmethod is competitive in respect to time consumption} When optimized for time consumption, \ourmethod outperforms all other baselines on Task Score (BertAA) while maintaining a time consumption less than the average of the baselines. A full analysis can be found in \cref{fig:time_analysis}. 

\section{Related Work}
\myparagraph{Stylometry} Stylometry, a field for statistically analyzing variations in writing styles, has long been used for authorship verification \cite{Calix2008StylometryFE, Fox2012StatisticalSA, jockersfederalist}. Consequently, employing stylometry as a means to assess writing style served as a logical extension in the task of authorship obfuscation.

\myparagraph{Stylometric Feature Approaches} Some approaches rely solely on stylometric features to create general numerical-based rules for obfuscation. For example, in a method submitted to the PAN 2016 Author Masking Shared Task by \citet{mansoorizadeh}, they substituted synonyms for the most frequently used terms in a text. Another method,  submitted to the same Shared Task was from \citet{stylo_method}, was more complex and used on a set of 500+ stylometric features such as average amount of words, word frequency, and punctuation. Based on these calculable attributes, the approach adjusted the text to bring the values closer to a pre-determined "average" (derived from a large training corpus). These approaches are often simple to implement, require no additional corpus, and may be used on any text. However, the rigidity of these rules often lead to incorrect grammar or non-fluent speech \cite{mahmood2019mutantx, pan2016_results}. 

\myparagraph{Model Based Approaches} Other approaches incorporate more flexibility by utilizing deep learning models. One of the most successful deep learning methods is the Support Vector Machine combined with Writeprint-Static\cite{amt_dataset}, which uses a collection of 500+ stylistic features from Writeprint \cite{writeprint} to construct a Support Vector Machine (SVM) model for authorship detection. It then uses this classifier as a guide in conjunction with a pattern disruption method. This framework inspired additional methods, such as Mutant-X \cite{mahmood2019mutantx}, a genetic algorithm that utilizes an internal classifier to iteratively "mutate" a sentence. At first this method used SVC or Random Forest architecture for the internal classifiers, but in later works reported to be more successful when an ensemble of classifiers was used \cite{avengers}. There has also been work which used variational autoencoder (VAE) network models to generate differentially private obfuscations \cite{weggenmann}. This was done using probabilistic encoders to do differentially private latent sampling. 

Another approach, which shares popularity with the task of paraphrasing, is round-trip machine translation using supervised language models. Initial implementations of this method relied on statistical machine translation techniques like Moses, as demonstrated in \citet{Keswani2016AuthorMT}. This approach involved translating text from English to German via French and then back to English. However, this method often produced nonsensical or inaccurate content \cite{ pan2016_results}. Fortunately, with the advancement of machine translation models, we have seen a significant increase in language quality \cite{altakrori-etal-2022-multifaceted}.

\myparagraph{Authorship Imitation Approach}
Although authorship imitation (or style transfer) is regarded as a distinct, separate task from authorship obfuscation, it can be used as an obfuscation strategy when the author's identity is known. For example, \citet{ Shetty2017A4NTAA} employ prior knowledge of the original authors' qualities such as age and gender to train a GAN-based model to generate content in multiple styles. For example, if the author is known to be an adult, this method would rewrite the section in a teenager’s tone. This strategy involves not only knowledge of the original author, but also a target style to shift to, making it a less general method for obfuscation. \citet{jones2022robert} also use a similar approach by training GPT2 models to successfully mimic blog or Twitter users to deceive authorship attribution models.

\section{Conclusion}
\label{sec:conclusion}
In this work, we introduced \ourmethod, a novel approach to user-controlled, inference-time authorship obfuscation which utilizes only small, open-source language models. This technique involves three key stages: keyword extraction, constrained diverse beam search, and filtering, offering users fine-grained control over the process and yielding personalized outcomes dependent on the user's needs. We showed experimentation on two diverse datasets, and demonstrated that \ourmethod outperformed over existing state-of-the-art methods in authorship obfuscation, while also showcasing its competitive performance against significantly larger models like GPT3.5. Our findings underscore the promise of \ourmethod as an effective strategy for authorship obfuscation, harnessing the capabilities of smaller, openly available models to achieve results on par with their larger counterparts.

\section{Limitations}
\label{sec:limitations}
\ourmethod has several limitations. First, for creation of the obfuscation candidates, we employ generations from a pre-trained language model. These models, however, have been known to add factually incorrect or hallucinatory information \cite{Ji2022SurveyOH}. Despite the fact that we have content-preserving filters, we have discovered that at times, additional information can bypass these filters and make it into the final obfuscation. 
 
Second, our approach is based on producing several candidates for each obfuscation. If the approach is employed at the sentence-level and the text is lengthy, it may take a long time to employ. Despite the fact that we demonstrated that our method works similarly with fewer generations, it is slower than traditional stylometric-based methods.

Lastly, the specific filtering techniques (e.g., NLI, CoLA) we used may carry biases into the eventual obfuscated texts. For example, CoLA might only be able to correctly filter standard, plain English language, but might not be as stable in certain dialects, which may exacerbate social injustice, e.g., correcting (whitewashing) African American English dialect. Users of this authorship obfuscation technique are strongly advised to examine the method for their specific text genre before deploying to ensure proper intended use.

Although we present our method with only beneficial use in mind, we acknowledge that the task of authorship obfuscation can be potentially dangerous in itself. First, it could be misused for anonymizing people's writing style for malicious intents, e.g., spamming or making hateful comments online without taking accountability for their actions. Also, these techniques could pose the risk of violating intellectual properties and rights when the creative work of authors is obscured to lose credits. We urge the user to think critically before using these types of methods. 

\section{Acknolwedgement} This research is based upon work supported in part by NSF DMS-2134012, DMS-2023166, CCF-2019844, and the Office of the Director of National Intelligence (ODNI)’s IARPA program via 2022-22072200003. The views and conclusions contained herein are those of the authors and should not be interpreted as  representing the official views of ODNI, IARPA, or the U.S. Government. \bibliography{custom.bib}
\bibliographystyle{acl_natbib}

\clearpage
\appendix
\onecolumn
{\huge Table of Contents: Appendix}\\
\\
In the appendix, we provide the following additional materials:
\begin{enumerate}[label={}]
    \item \textbf{\cref{appx:additional_experiements}}: Additional Experiments 
    \begin{itemize}
        \item \cref{appx:diversity_exp}: Impact of Diversity in Beam Search
        \item \cref{appx:extra_human_eval}: Human Evaluation without CoLA Threshold
        \item \cref{appx:keyword_extraction_comparison}: Comparing Keyword 
        \item \cref{appx:lightweight_exp}: \ourmethod with Smaller Beam Width
        \item \cref{appx:metric_comparison_all}: Comparison of all Automatic Evaluations
        \item \cref{appx:threshold_changing_exp}: Affect of NLI/CoLA Threshold on Performance
        \item \cref{appx:perplex}: Average Perplexity of Text
        \end{itemize}
    \item \textbf{\cref{appx:style_transfer}}: Style Transfer as Authorship Obfuscation Method
    \item \textbf{\cref{appx:threat_model}}: Adversarial Threat Model for Evaluation
    \item \textbf{\cref{appx:additional_example}}: Additional Qualitative Example for Comparison of Methods
    \item \textbf{\cref{appx:time_consumption}}: Time Consumption Analysis
    \item \textbf{\cref{appx:compare_author_tasks}}: Compare Similar Authorship Tasks
    \item \textbf{\cref{appx:exp_details}}: Experimentation Details
    \item \textbf{\cref{appx:algorithms}}: Algorithm for Constrained Diverse Beam Search \cdbs
\end{enumerate} \clearpage
\twocolumn
\section{Additional Experiments}\label{appx:additional_experiements}

\subsection{Impact of Combining Diverse Beam Search with Constrained Beam Search}\label{appx:diversity_exp} In order to explore the impact of combining Diverse Beam Search \cite{vijayakumar2018diverse} and Constrained Beam Search \cite{post2018fast} for authorship obfuscation, we calculated the automatic evaluation metrics on generations produced using \ourmethod with and without the Diverse Beam Search for the AMT datasets. Results are shown in \cref{tab:amt_ablation_diversity_class}. On average, there is about an $6\%$ increase in the Drop Rate, as well as an average $32\%$ increase in generations that pass the NLI and CoLA thresholds, with little change to the NLI and CoLA scores. As expected, adding the diversity penality successfully encourages a higher diversity of generations between beams resulting in a more diverse pool of generation candidates.

\subsection{Human Evaluation for \ourmethod+Stylo without CoLA Threshold}\label{appx:extra_human_eval}
\begin{figure}[h!]
\begin{flushleft}
\includegraphics[width=0.5\textwidth]{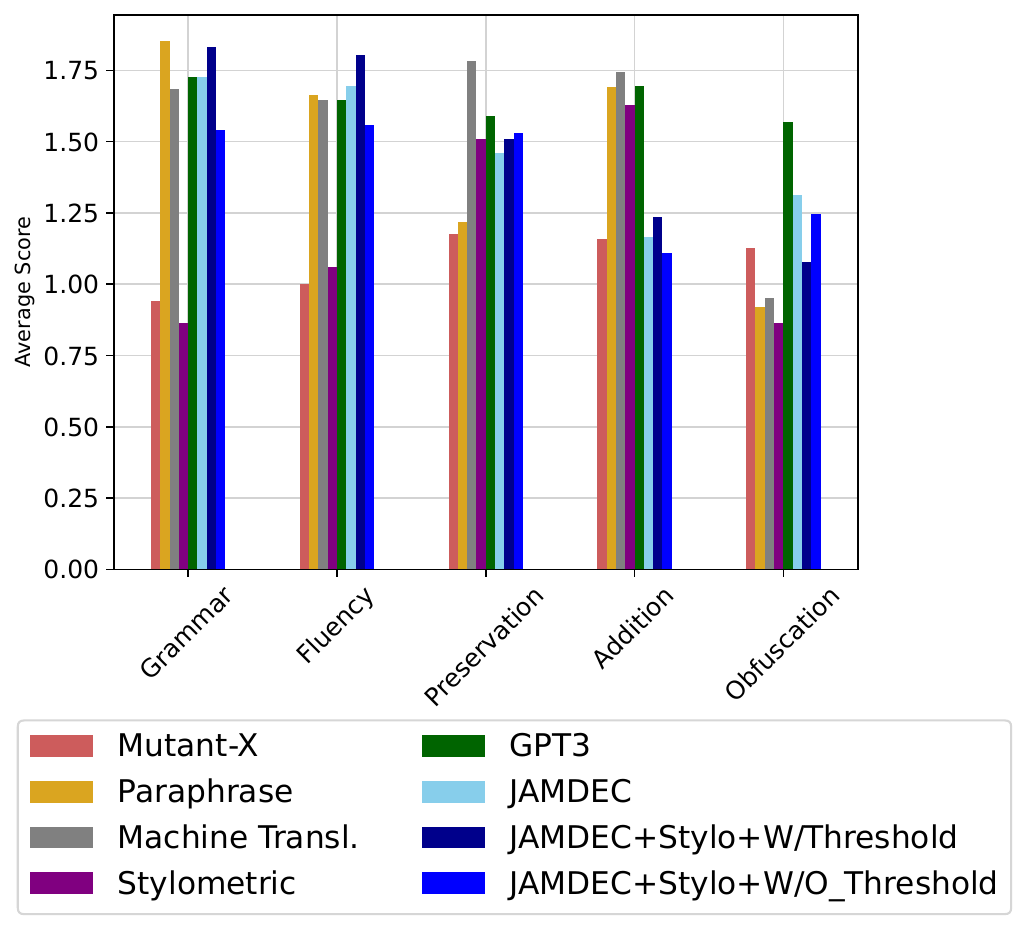}
\caption{Human Evaluation on 102 random samples from AMT-3. We include two versions of \ourmethod+Stylo, the original that uses a final CoLA threshold (\ourmethod+Stylo+W/\_Threshold) and one that does not use this threshold (\ourmethod+Stylo+W/O\_Threshold). }
\label{fig:human_eval_abl}
\end{flushleft}
\end{figure}

We ran an additional human evaluation on a third variant of \ourmethod, which is identical to \ourmethod+Stylo except it does not include the final CoLA threshold on sentences produced using the stylometric-based obfuscation method.  Without this final threshold, each sentence obfuscated using the stylometric-based method was included in the final text, meaning all sentences of the text were changed and no original text was used. For simplicity, we distinguish these methods as \ourmethod+Stylo+W/Threshold and \ourmethod+Stylo+W/O\_Threshold. \cref{fig:human_eval_abl} compares these results to the results shown earlier in \cref{sec:expt}. We observe an overall increase in Obfuscation of $9\%$ compared to \ourmethod+Stylo+W/Threshold, making it higher than all task-specific methods (but still slightly below \ourmethod). However, it did have a decrease of $15\%$ and $13\%$ in
Grammar and Fluency, respectively. The obfuscated text in \ourmethod+Stylo+W/O\_Threshold only differs from \ourmethod+Stylo+W/Threshold for sentences that were altered by the stylometric-based obfuscation method but did not pass the CoLA threshold. Therefore, it logically follows that including these sentences leads to a decrease in Grammar and Fluency. It also follows that these changes would add to a slight increase in obfuscation, compared to text which includes some of the original sentences.
\begin{table}[t]\centering
    \resizebox{.5\textwidth}{!}{
    \begin{tabular}{ llcc }\toprule
        
        {\textbf{Dataset}}&{\textbf{Metric}} & \textbf{W/ Diversity} & \textbf{W/O Diversity}\\  
        \midrule
\amt{3}&Drop Rate (ENS)&\textbf{0.11}&	0.01\\
 &Drop Rate (BertAA)&  	0.04&	\textbf{0.08} \\
   &NLI&  	0.75&	\textbf{0.87 }\\
    &CoLA&  	0.85&	\textbf{0.86} \\
    &Average Gen. & \textbf{0.52}& 0.16\\
        \midrule
\amt{5} &Drop Rate (ENS)&  \textbf{0.10}	&	\textbf{0.10}\\
&Drop Rate (BertAA)& \textbf{0.14}&	0.01\\
   &NLI&  	0.76&	\textbf{0.87} \\
    &CoLA&  	0.85 &	\textbf{0.87 }\\
        &Average Gen. & \textbf{0.48}& 0.16\\
        \midrule
\amt{10}&Drop Rate (ENS)& \textbf{0.44}&	0.25\\
 &Drop Rate (BertAA)& -0.03&\textbf{ 0.00}	\\
    &NLI&  	0.79&	\textbf{0.85} \\
    &CoLA&  	0.78&	\textbf{0.85 }\\
            &Average Gen. & \textbf{0.47}& 0.18\\
        \bottomrule
    \end{tabular}
    }
    \caption{The results of the Drop Rate, NLI, and CoLA scores using \method with the same parameters both with and without including a diversity penalty with Constrained Beam Search. We also present the average generations that pass the NLI/CoLA threshold ("Average Gen.") for each method.}
    \label{tab:amt_ablation_diversity_class}
\end{table}

\subsection{Comparing Keyword Extractors: Word Embedding Methods vs. Likelihood Methods}\label{appx:keyword_extraction_comparison}
\begin{figure*}[h!]
\begin{flushleft}
\includegraphics[width=1\textwidth]{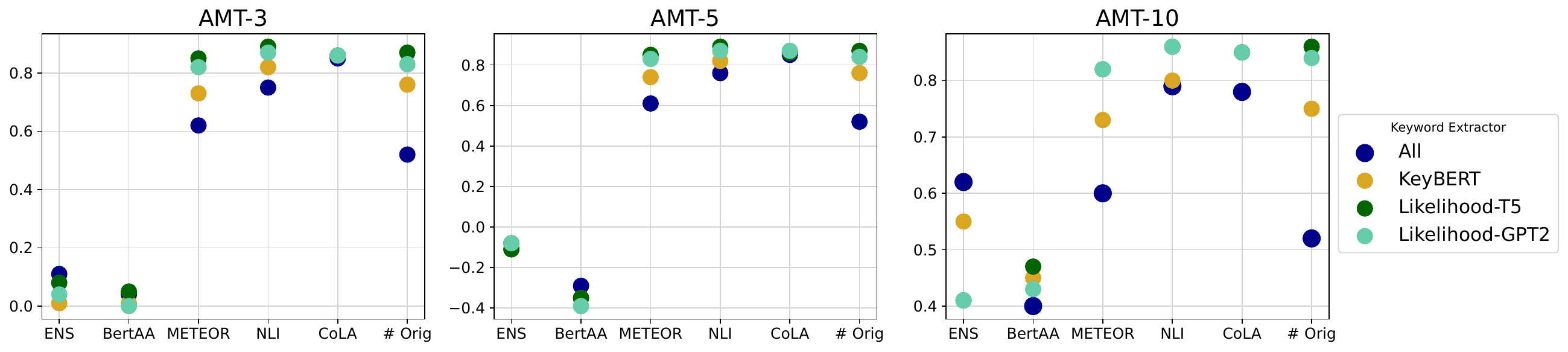}
\caption{Comparing the obfuscation (Drop Rate - ENS and BertAA), content preservation (NLI), and language quality (CoLA) using each keyword extraction method individually (KeyBERT, Likelihood-T5, Likelihood-GPT2, and all three together (All) for \amt{3}, \amt{5}, and \amt{10}.}
\label{fig:comparison_keyword_extractor}
\end{flushleft}
\end{figure*}
In \cref{sect:method} we introduced a new framework for keyword extraction which uses likelihoods of next token prediction from language models instead of word embeddings. Using this framework, we developed two keyword extraction methods; one using T5 and infilling (Likelihood-T5), and the other using GPT2 with an autoregressive (left to right) generation (Likelihood-GPT2). We hypothesized that these likelihood-based keyword extraction methods would highlight keywords that would increase the ability of a downstream model to generate text that preserves the original meaning. In \cref{fig:comparison_keyword_extractor} we show the results of the automatic evaluations of authorship obfuscation using generations created either with only KeyBERT, only Likelihood-T5, only Likelihood-GPT2, or all three (as we did in our experiments). For \amt{3} and \amt{5}, the likelihood-based keyword extraction have higher overall evaluations' metrics than the embedding-based (KeyBERT). However, in \amt{10}, the KeyBERT performs on average $\sim 10\%$ higher than both the likelihood method in Drop Rate (ENS), but is on average  $6\%$ lower in NLI. Overall, the combined method (using all three keyword extraction) has the highest Drop Rate overall and lowest number of original sentences used. Examples of keywords selected by each method can be reviewed in \cref{tab:keyword_examples}.

\begin{table}[h!]
\centering
\resizebox{.5\textwidth}{!}{
\begin{tabular}{ll}
 \hline
 \rowcolor{platinum}Original Sentence& "I stated that the body needs a specific amount of time to \\
 \rowcolor{platinum}& transfer calcium from locations in the body to the fracture."  \\ 
 \hline
  \textbf{Keyword Extractor} & \textbf{Keywords} \\  
 \midrule
 
 KeyBERT &  ["stated", "body", "needs", "specific", "time", "transfer", "calcium"]\\
 \midrule
 Likelihood-T5 & ["that", "the", "body", "of", "time", "to", "from", "location"]\\
 \midrule
 Likelihood-GPT2 & ["stated", "needs", "of", "transfer", "calcium"]\\
 \hline
\end{tabular}
}
\caption{Examples of keywords extracted by each method; KeyBERT, Likelihood-T5, and Likelihood-GPT2.}
\label{tab:keyword_examples}
\end{table}

\subsection{\ourmethod with Smaller Beam Widths (Less Generations)}\label{appx:lightweight_exp} 
We repeated the AMT-3 experiment using a lightweight \ourmethod with a smaller beam width ($20$) and discovered that it performs slightly better on almost all metrics than \ourmethod with a larger beam width ($50$) (results in \cref{tab:amt_ablation_lightweight}). This appeared odd at first, until we looked at the quantity of sentences that had generations which passed the NLI and CoLA filter. When we reduce the beam width (and hence the number of overall generations produced), we find a significant decrease in the number of generations that pass the thresholds. For example, in the lightweight version (beam width = 20), only $20\%$ of the generations pass the threshold, implying that $80\%$ of the sentences reverted to the original sentence. Although changing only $20\%$ of the sentences is sufficient to trick the classifiers (seen in the almost matching Drop Rate), it may not be sufficient in human-evaluation.

\begin{table}[t]\centering
    \resizebox{.5\textwidth}{!}{
    \begin{tabular}{ lcccc }\toprule
          {\textbf{Metric}}  & \multicolumn{1}{c}{\textbf{\ourmethod}} & \multicolumn{1}{c}{\textbf{\ourmethod (Lightweight)}}\\
        \midrule
        Drop Rate (ENS)  & 0.11&\textbf{0.12}\\
    Drop Rate (BertAA)  &\textbf{0.04}& \textbf{0.04}\\
        METEOR &0.62& \textbf{0.78}\\
       NLI &0.81&\textbf{0.82}\\
      CoLA  &0.79& \textbf{0.83}\\
      Average Gen. & \textbf{0.63}& 0.42\\
         \rowcolor{platinum}   Task Score (ENS) &0.57& \textbf{0.59}\\
          \rowcolor{platinum}        Task Score (BertAA)&0.55& \textbf{0.56} \\
        \bottomrule
    \end{tabular}
    }
    \caption{The results of the automatic evaluation scores for AMT-3 using \method with different beam widths/generations per beam search (50 vs. 20). We also present the average generations that pass the NLI/CoLA threshold ("Average Gen.") for each method. }
    \label{tab:amt_ablation_lightweight}
\end{table}
 
\subsection{Drop Rate vs. NLI vs. CoLA for All Methods}\label{appx:metric_comparison_all} 
\begin{figure*}
\begin{subfigure}[b]{0.5\textwidth}
    \includegraphics[scale=.3]{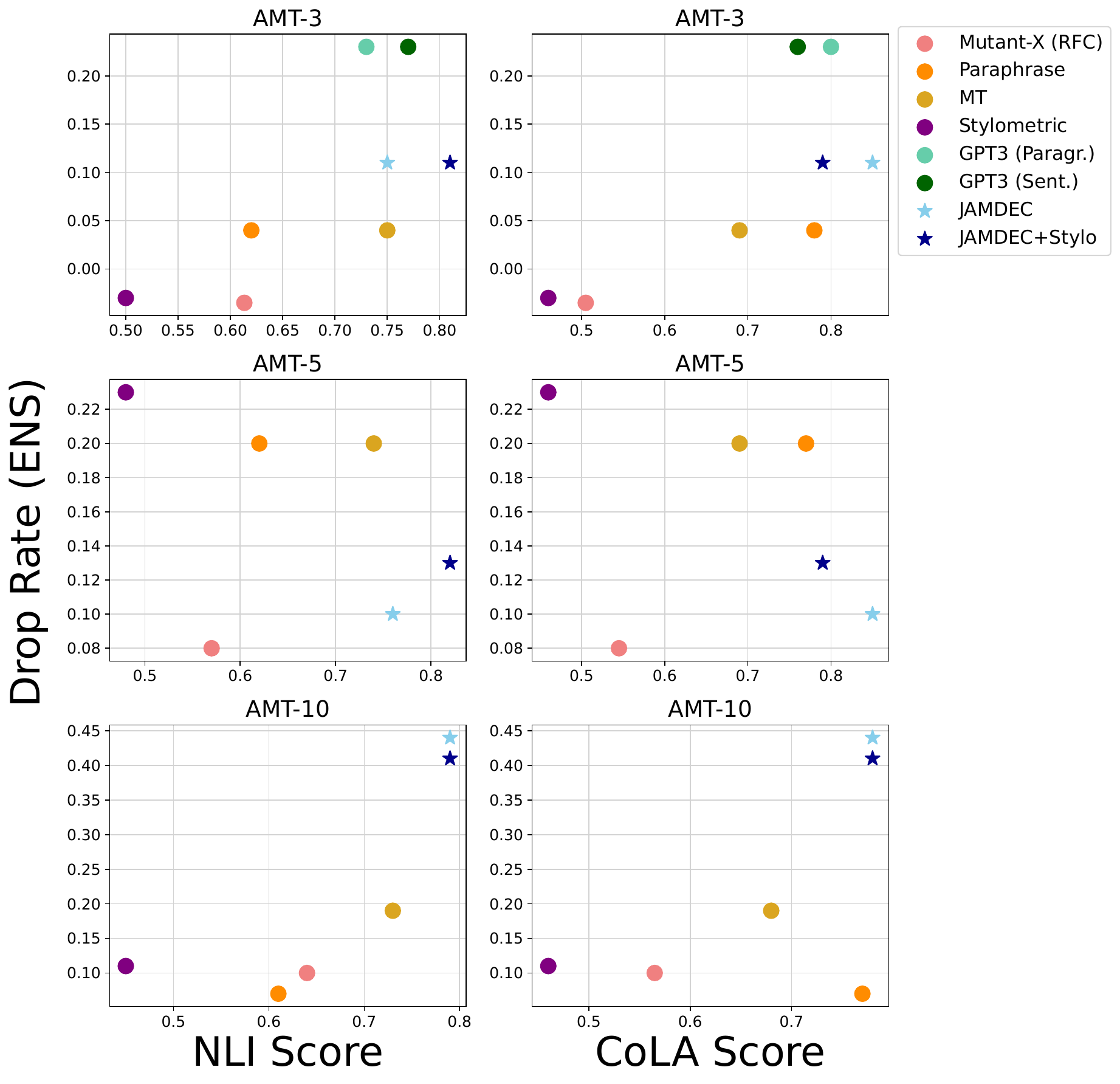}
    \caption{AMT Datasets}
\end{subfigure}
\begin{subfigure}[b]{0.5\textwidth}
    \includegraphics[scale=.3]{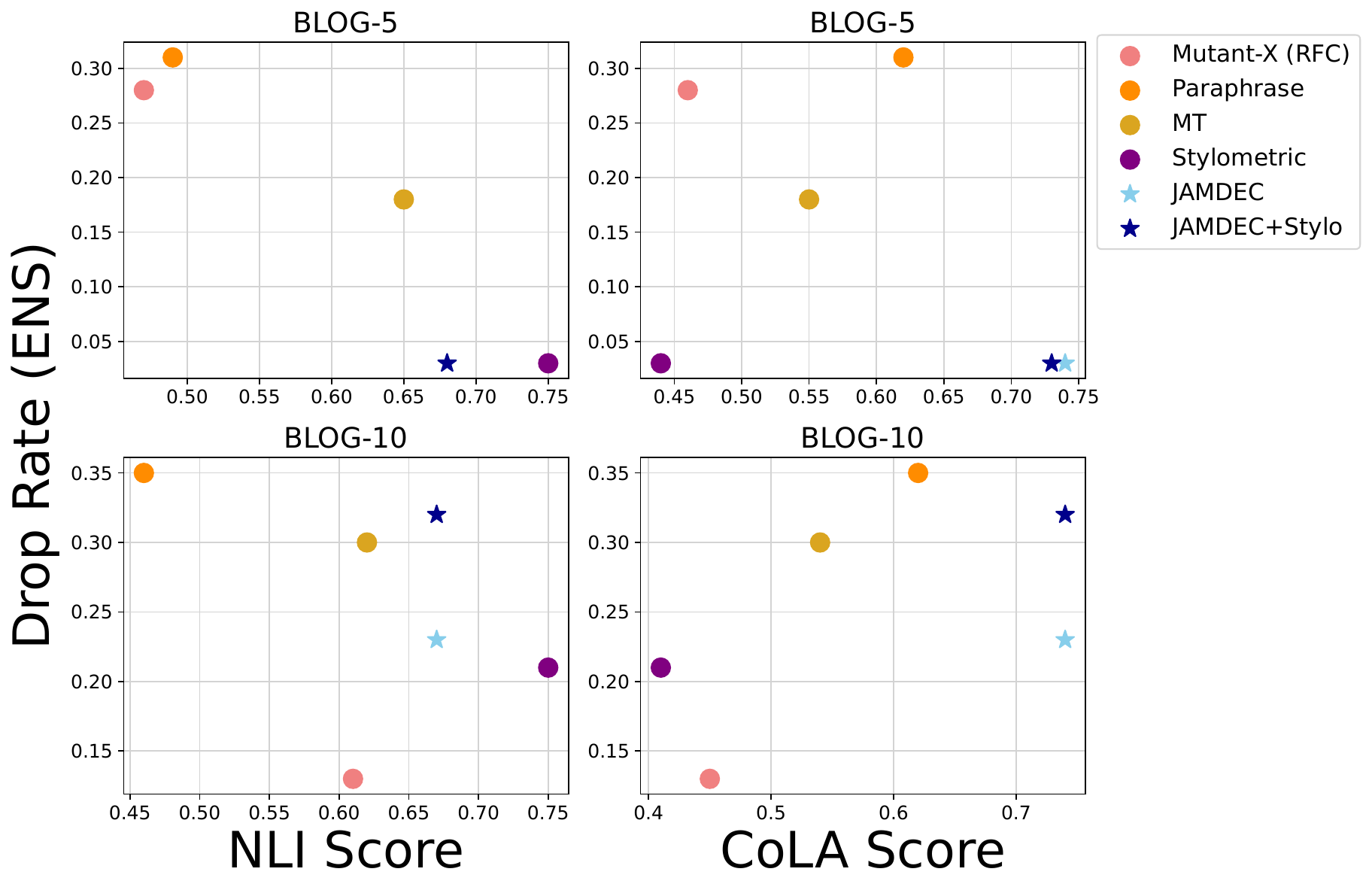}
    \caption{BLOG Datasets}
\end{subfigure}
\caption{Highlighting the tradeoff between obfuscation (Drop Rate (ENS)), content preservation (NLI), and language quality (CoLA) of each method for all datasets. The dotted line indicates the trend through all methods.}
\label{fig:nli_cola_tradeoff_all_data_all}
\end{figure*}

A successful authorship obfuscation method should score high in Drop Rate, NLI, and CoLA, however we observe that the current methods tend to have a trade-off in their abilities. To further analyze this tradeoff, in \cref{fig:nli_cola_tradeoff_all_data_all} we graph the Drop Rate (ENS) versus the NLI and CoLA separately for all datasets. Using our definition of a successful method, we want to have a method that lies in the top right of both graphs. We observe that for both datasets (AMT and BLOG), authors 3 and 10, \ourmethod has both a higher Drop Rate and a high NLI and CoLA compared to all other small model methods. However, we do see it perform a bit worse for the 5 authors datasets, where Machine Translation is a bit higher in Drop Rate and close in NLI. 

\subsection{Comparing Drop Rate, NLI, and CoLA for \ourmethod as the NLI/CoLA Thresholds Change}\label{appx:threshold_changing_exp}
\begin{figure*}[t!]
\begin{flushleft}
\includegraphics[width=1\textwidth]{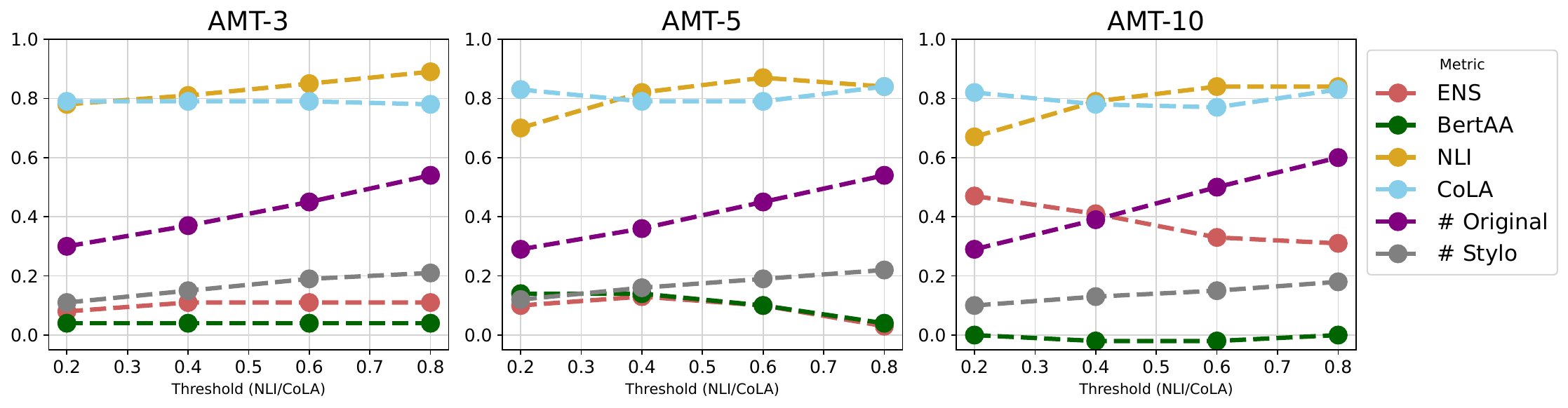}
\caption{Highlighting the change in obfuscation (Drop Rate - ENS and BertAA), content preservation (NLI), and language quality (CoLA) for the \ourmethod+Stylo method as we increase the NLI/CoLA threshold for AMT-3.}
\label{fig:nli_cola_tradeoff_all_data}
\end{flushleft}
\end{figure*}

\ourmethod is designed to be user-adaptive, having flexible hyperparameters that can adjust to the needs of the specific task. Two of these hyperparameters are the base NLI and CoLA thresholds used in the filtering stage. We experimented with scaling these hyperparameters from $0.2$ to $0.8$, using the \ourmethod+Stylo method. For simplicity, we make the NLI and CoLA threshold equal in each experimentation, and use a constant final CoLA threshold of $0.7$. \cref{fig:nli_cola_tradeoff_all_data} shows the results for the AMT datasets. In general, as we increase the NLI and CoLA Thresholds (making it harder for generation candidates to pass) we see an obvious increase in NLI of $\sim 15\%$, a steady score of CoLA, and a mixed result for the Drop Rate depending on the number of authors. In fact, we see a slight increase in both Drop Rates for AMT-3 and a slight decrease in AMT-5 and AMT-10. Since the number of original sentences used increases as the threshold increases (higher thresholds means less generations pass the thresholds), we would expect Drop Rate to decrease (as it did for AMT-3). Therefore, this behavior (especially by ENS) is an indication that it might be relying on an artifact for its classification. This encourages the use of human evaluation as added evaluation for this task. 

\subsection{Perplexity of Generations}\label{appx:perplex} In our main experimentation we do not use perplexity and instead use the CoLA score. The reason we opted for CoLA over perplexity is that it has a fixed range $[0,1]$ and can therefore be compared across text length, topic, and style type (formal/informal). Due to the unbounded nature of perplexity, it is an unreliable metric to use by itself \cite{wang2023perplexity}.

However, want to provide these metrics. We have used a Llama2-7B model \cite{touvron2023llama} to calculate perplexity over a text (normalized to the length of text). We choose Llama2-7B since it is from a different family of models used in our experimentation, to reduce any model architecture bias. Then, we calculate the ratio of the perplexity of the obfuscated text to the perplexity of the original (human) text. Again, we use this ratio to have a standard comparison across methods. Results can be seen in \cref{tab:perplexity}. Similar to CoLA, we see that JAMBDEC outperforms over all other methods on perplexity (ratio closes to 1) on almost all datasets.
\begin{table*}[]
\begin{tabular}{lllll}
\hline
\textbf{Dataset}& \textbf{Method}& \textbf{Original Perplexity} & \textbf{Predicted Perplexity} & \textbf{ Ratio}       \\ \hline
{\color[HTML]{000000} }   & {\color[HTML]{000000} Mutant-X (ENS)}      & {\color[HTML]{000000} 8.25}                & {\color[HTML]{000000} 29.52}                & {\color[HTML]{000000} 3.77}          \\
{\color[HTML]{000000} }        & {\color[HTML]{000000} Mutant-X (SVC)}      & {\color[HTML]{000000} 8.25}                & {\color[HTML]{000000} 27.6}                 & {\color[HTML]{000000} 3.56}          \\
{\color[HTML]{000000} }        & {\color[HTML]{000000} Paraphrase}          & {\color[HTML]{000000} 8.25}                & {\color[HTML]{000000} 9.8}                  & {\color[HTML]{000000} 1.23}          \\
{\color[HTML]{000000} }        & {\color[HTML]{000000} Machine Translation} & {\color[HTML]{000000} 8.25}                & {\color[HTML]{000000} 12.93}                & {\color[HTML]{000000} 1.64}          \\
{\color[HTML]{000000} AMT-3}        & {\color[HTML]{000000} Stylometric}         & {\color[HTML]{000000} 8.25}                & {\color[HTML]{000000} 24.17}                & {\color[HTML]{000000} 2.95}          \\
{\color[HTML]{000000} }        & {\color[HTML]{000000} \ourmethod (w/o stylo)} & {\color[HTML]{000000} 8.25}                & {\color[HTML]{000000} 7.29}                 & {\color[HTML]{000000} \textit{0.92}} \\
{\color[HTML]{000000} }        & {\color[HTML]{000000} \ourmethod (w stylo)}   & {\color[HTML]{000000} 8.25}                & {\color[HTML]{000000} 8.05}                 & {\color[HTML]{000000} \textbf{1.02}} \\
{\color[HTML]{000000} }        & {\color[HTML]{000000} GPT3 (Sentence)}     & {\color[HTML]{000000} 8.25}                & {\color[HTML]{000000} 13.3}                 & {\color[HTML]{000000} 1.7}           \\
{\color[HTML]{000000} }        & {\color[HTML]{000000} GPT3 (Paragraph)}    & {\color[HTML]{000000} 8.25}                & {\color[HTML]{000000} 9.96}                 & {\color[HTML]{000000} 1.23}          \\ \hline
{\color[HTML]{000000} }   & {\color[HTML]{000000} Mutant-X (ENS)}      & {\color[HTML]{000000} 8.42}                & {\color[HTML]{000000} 285.92}               & {\color[HTML]{000000} 34.08}         \\
{\color[HTML]{000000} }        & {\color[HTML]{000000} Mutant-X (SVC)}      & {\color[HTML]{000000} 8.42}                & {\color[HTML]{000000} 923.09}               & {\color[HTML]{000000} 117.55}        \\
{\color[HTML]{000000} }        & {\color[HTML]{000000} Paraphrase}          & {\color[HTML]{000000} 8.42}                & {\color[HTML]{000000} 11.95}                & {\color[HTML]{000000} \textit{1.49}} \\
{\color[HTML]{000000} AMT-5}        & {\color[HTML]{000000} Machine Translation} & {\color[HTML]{000000} 8.42}                & {\color[HTML]{000000} 13.41}                & {\color[HTML]{000000} 1.66}          \\
{\color[HTML]{000000} }        & {\color[HTML]{000000} Stylometric}         & {\color[HTML]{000000} 8.42}                & {\color[HTML]{000000} 25.81}                & {\color[HTML]{000000} 3.09}          \\
{\color[HTML]{000000} }        & {\color[HTML]{000000} \ourmethod (w/o stylo)} & {\color[HTML]{000000} 8.42}                & {\color[HTML]{000000} 7.3}                  & {\color[HTML]{000000} \textbf{0.9}}  \\
{\color[HTML]{000000} }        & {\color[HTML]{000000} \ourmethod (w stylo)}   & {\color[HTML]{000000} 8.42}                & {\color[HTML]{000000} 37.56}                & {\color[HTML]{000000} 4.44}          \\ \hline
{\color[HTML]{000000} }  & {\color[HTML]{000000} Mutant-X (ENS)}      & {\color[HTML]{000000} 9.07}                & {\color[HTML]{000000} 25.96}                & {\color[HTML]{000000} 3.08}          \\
{\color[HTML]{000000} }        & {\color[HTML]{000000} Mutant-X (SVC)}      & {\color[HTML]{000000} 9.07}                & {\color[HTML]{000000} 23.51}                & {\color[HTML]{000000} 2.77}          \\
{\color[HTML]{000000} }        & {\color[HTML]{000000} Paraphrase}          & {\color[HTML]{000000} 9.07}                & {\color[HTML]{000000} 10.02}                & {\color[HTML]{000000} \textit{1.2}}  \\
{\color[HTML]{000000} AMT-10}        & {\color[HTML]{000000} Machine Translation} & {\color[HTML]{000000} 9.07}                & {\color[HTML]{000000} 15.16}                & {\color[HTML]{000000} 1.79}          \\
{\color[HTML]{000000} }        & {\color[HTML]{000000} Stylometric}         & {\color[HTML]{000000} 9.07}                & {\color[HTML]{000000} 26.24}                & {\color[HTML]{000000} 2.88}          \\
{\color[HTML]{000000} }        & {\color[HTML]{000000} \ourmethod (w/o stylo)} & {\color[HTML]{000000} 9.07}                & {\color[HTML]{000000} 7.52}                 & {\color[HTML]{000000} \textbf{0.9}}  \\
{\color[HTML]{000000} }        & {\color[HTML]{000000} \ourmethod (w stylo)}   & {\color[HTML]{000000} 9.07}                & {\color[HTML]{000000} 34.65}                & {\color[HTML]{000000} 3.86}          \\ \hline
{\color[HTML]{000000} }  & {\color[HTML]{000000} Mutant-X (ENS)}      & {\color[HTML]{000000} 22.82}               & {\color[HTML]{000000} 89.53}                & {\color[HTML]{000000} 5.24}          \\
{\color[HTML]{000000} }        & {\color[HTML]{000000} Mutant-X (SVC)}      & {\color[HTML]{000000} 22.82}               & {\color[HTML]{000000} 55.04}                & {\color[HTML]{000000} 3.73}          \\
{\color[HTML]{000000} }        & {\color[HTML]{000000} Paraphrase}          & {\color[HTML]{000000} 22.82}               & {\color[HTML]{000000} 22.27}                & {\color[HTML]{000000} \textbf{1.39}} \\
{\color[HTML]{000000} BLOG-5}        & {\color[HTML]{000000} Machine Translation} & {\color[HTML]{000000} 22.82}               & {\color[HTML]{000000} 42.08}                & {\color[HTML]{000000} 2.73}          \\
{\color[HTML]{000000} }        & {\color[HTML]{000000} Stylometric}         & {\color[HTML]{000000} 22.82}               & {\color[HTML]{000000} 47.18}                & {\color[HTML]{000000} 2.5}           \\
{\color[HTML]{000000} }        & {\color[HTML]{000000} \ourmethod (w/o stylo)} & {\color[HTML]{000000} 22.82}               & {\color[HTML]{000000} 23.79}                & {\color[HTML]{000000} \textit{1.7}}  \\
{\color[HTML]{000000} }        & {\color[HTML]{000000} \ourmethod (w stylo)}   & {\color[HTML]{000000} 22.82}               & {\color[HTML]{000000} 24.44}                & {\color[HTML]{000000} 1.75}          \\ \hline
{\color[HTML]{000000} } & {\color[HTML]{000000} Mutant-X (ENS)}      & {\color[HTML]{000000} 19.55}               & {\color[HTML]{000000} 452.56}               & {\color[HTML]{000000} 32.25}         \\
{\color[HTML]{000000} }        & {\color[HTML]{000000} Mutant-X (SVC)}      & {\color[HTML]{000000} 19.55}               & {\color[HTML]{000000} 47.82}                & {\color[HTML]{000000} 3.58}          \\
{\color[HTML]{000000} }        & {\color[HTML]{000000} Paraphrase}          & {\color[HTML]{000000} 19.55}               & {\color[HTML]{000000} 20.82}                & {\color[HTML]{000000} 1.8}           \\
{\color[HTML]{000000} BLOG-10}        & {\color[HTML]{000000} Machine Translation} & {\color[HTML]{000000} 19.55}               & {\color[HTML]{000000} 42.93}                & {\color[HTML]{000000} 3.16}          \\
{\color[HTML]{000000} }        & {\color[HTML]{000000} Stylometric}         & {\color[HTML]{000000} 19.55}               & {\color[HTML]{000000} 45.63}                & {\color[HTML]{000000} 2.74}          \\
{\color[HTML]{000000} }        & {\color[HTML]{000000} \ourmethod (w/o stylo)} & {\color[HTML]{000000} 19.55}               & {\color[HTML]{000000} 19.17}                & {\color[HTML]{000000} \textbf{1.4}}  \\
{\color[HTML]{000000} }        & {\color[HTML]{000000} \ourmethod (w stylo)}   & {\color[HTML]{000000} 19.55}               & {\color[HTML]{000000} 19.72}                & {\color[HTML]{000000} \textit{1.44}} \\\hline
\end{tabular}
\caption{Perplexity of Experiments in main text. The ratio closes to 1 (better fluency) is \textbf{bolded} and the second best is \textit{italized}.} 
\label{tab:perplexity}
\end{table*} \section{Style Transfer as Authorship Obfuscation Method}\label{appx:style_transfer}
As we mentioned, the task of style transfer mainly differs from the task of authorship obfuscation by its goal of a specific, fixed target style. For this reason, there seems to be many subclasses of style transfer tasks center on a specific aspect of style (specific authors, such as characters from the TV show Friends \cite{Tikhonova2021StyleTI}, aspect of authors, such as gender \cite{Tokpo2022TextST}, formality of style \cite{Chen2022StyleTA}, etc.). This makes it hard to be a main baseline for authorship obfuscation, as there is not a specific, unbiased method or target style to choose. However, we still were curious how it would compare to \ourmethod. Therefore, we have included an additional experimentation which compares two targeted styles with \ourmethod on the task of authorship obfuscation. 

We use the Style Transfer via Paraphrasing or STRAP, a clever method which first employs paraphrasing using one LLM finetuned on a supervised paraphrasing task and then applies a specific style using another LLM finetuned on the specific style \cite{krishna-etal-2020-reformulating}. We use two types of target styles; Shakespeare and Formal writing. 
The results are shown in \cref{tab:style_transfer}. Here we observe that \ourmethod consistently achieves a higher Drop Rate while better preserving content and maintaining fluency. Notice that comparing fluency using the style transfer baseline to Shakespearean style might not be entirely fair, as Old English has different grammar rules. This highlights the limitations of using the style transfer method for authorship obfuscation, given the lack of a specific, unbiased target style to select.

\begin{table*}[t]\centering
    \begin{tabular}{ llccc }\toprule
        
        {\textbf{Dataset}}&{\textbf{Metric}} & \textbf{Shakespeare} & \textbf{Formal} & \textbf{\ourmethod}\\  
        \midrule
\amt{3}&Drop Rate (ENS)&0.0&	0.0 & \textbf{0.11}\\
 &Drop Rate (BertAA)&  \textbf{0.04}	&	\textbf{0.04}&\textbf{0.04} \\
   &NLI&  	0.19&	0.25&  \textbf{0.75}\\
    &CoLA&  	0.47&	0.69 & \textbf{0.85} \\
        \midrule
\amt{5} &Drop Rate (ENS)& \textbf{0.20}	&	\textbf{0.20} & 0.13\\
&Drop Rate (BertAA)& -0.06&	-0.06&\textbf{0.14}\\
   &NLI&  	0.23&	0.26 & \textbf{0.76}\\
    &CoLA&  	0.49 &	0.69 &\textbf{0.85}\\
        \midrule
\amt{10}&Drop Rate(ENS)& 0.33&	0.23 &\textbf{0.41}\\
 &Drop Rate (BertAA)& \textbf{-0.02}&	-.04 &\textbf{-0.02}\\
    &NLI&  	0.19&	0.26 &\textbf{0.79} \\
    &CoLA&  	0.47&	0.67 & \textbf{0.78}\\
        \bottomrule
    \end{tabular}
    \caption{Results from the automatic evaluation for \ourmethod and style transfer methods on AMT dataset.}
    \label{tab:style_transfer}
\end{table*} \section{Threat Model as Evaluation}\label{appx:threat_model}
\begin{table}[t]\centering
    \resizebox{.5\textwidth}{!}{
    \begin{tabular}{ lcccc }\toprule
          {\textbf{Method}}  & \multicolumn{1}{c}{\textbf{Threat Model (Orig + Obf)}} & \multicolumn{1}{c}{\textbf{Threat Model (Obf)}}\\

        \midrule
        Mutant-X (ENS)  & 0.0&\textbf{0.03}\\
    Mutant-X (RFC) &0.0& 0.00\\
        Paraphrase &0.0& -0.03\\
       Machine Transl. &\textbf{0.4}&0.00\\
      Stylometric &0.00& -0.07\\
      \ourmethod & \textbf{0.04}& -0.03\\
      \hline
     \textbf{ Accuracy} & &\\
      Train & 1.0 & 1.0\\
      Test  & 0.93& 0.96\\
        \bottomrule
    \end{tabular}
    }
    \caption{Drop Rate for \method and other baseline methods on AMT-3 dataset. The threat models are used to assess the Drop Rate (average obfuscated text).}
    \label{tab:threat_model}
\end{table}
 In our main evaluation, we use simple authorship attribution models, which do not have knowledge of obfuscations. However, current work in authorship attribution has shown that the use of adversarial threat models (models that are trained with obfuscation) can better evade the attacks of authorship obfuscation \cite{Zhai2022AGH}. Therefore, we include evaluation using stronger threat models on the AMT-3 dataset.

\cref{tab:threat_model} shows results of evaluation of all methods using two threat models. The first, Threat Model (Orig + Obf),  is trained using both the original text and the obfuscated text from all methods shown. The second, Threat Model (Obf), is only trained using the same obfuscated text but no original text. It has been shown in previous works that threat models trained only on obfuscated text have higher accuracy \cite{Zhai2022AGH}, which is also seen in the models we train. 
Using these models, we see that \ourmethod has the highest Drop Rate under the first thread model and third highest under the second thread model. However, as mentioned before, the Drop Rate is only one criterion for the task evaluation of authorship obfuscation. It should be noted, that Mutant-X and Machine Translation (which are the only method which scores much higher than \ourmethod under the second threat model) scores much lower in language quality and content preservation than \ourmethod, as shown in Table \ref{tab:amt_class_metric}.

\section{Additional Example of Obfusction}\label{appx:additional_example}

\begin{figure}[h!]
\begin{flushleft}
\includegraphics[width=.5\textwidth]{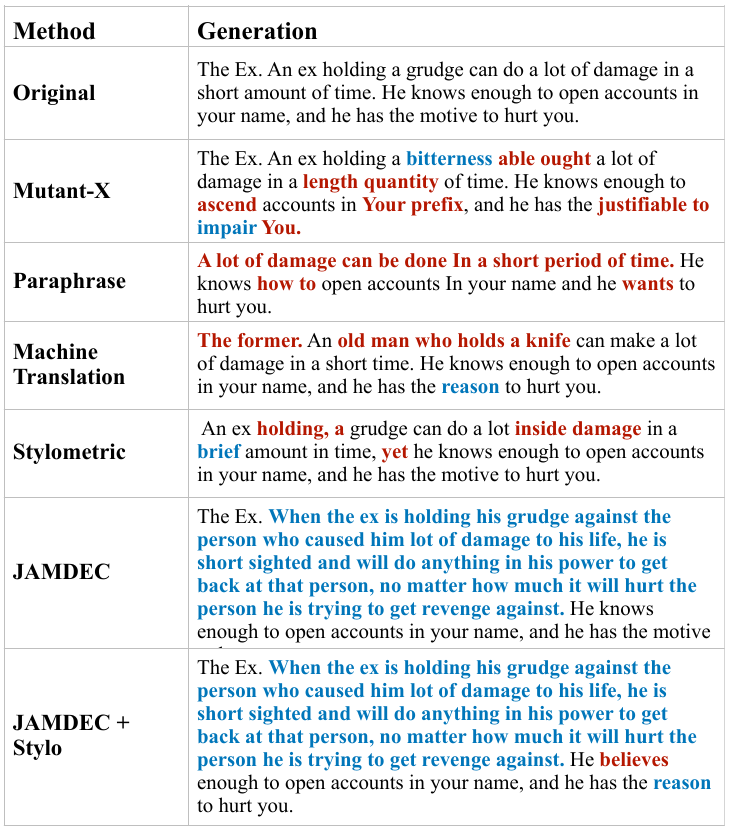}
\caption{Qualitative examples of obfuscated text created by each method. The sentences are taken from the AMT-3 dataset. Changes to the original are outline in \textcolor{teal}{\textbf{blue}} (correct grammatically and in context) and \textcolor{Maroon}{\textbf{red}} (incorrect grammatically or in context).}
\label{fig:qual_examples2}
\end{flushleft}
\end{figure}

In \cref{fig:qual_examples2} we include a second qualitative comparison of \ourmethod and the other baseline methods. We notice that the obfuscated text produced by baseline methods like Mutant-X, Paraphrase, and Machine Translation has much lower language quality compared to \ourmethod. Such low-quality text might make it easier to deceive an automatic classifier, but it fails to meet the other objectives of authorship obfuscation: preserving the quality and content of the original text. We also observe that Paraphrase and Machine Translation make only minor modifications to the original text. While this aids content preservation, it's ineffective for authorship concealment.

Also, we provide a few examples of GPT3.5 generation in \cref{table: gpt3.5_examples}, with the first being the same examples from \cref{fig:qual_examples2} in our paper. From qualitative analysis, we found that most generations from GPT3.5 fell within two techniques: paraphrasing and stylometric (mainly replacing words with synonyms). Either the generation was a short description (lacking some content preservation) or it was minimally changed (only swapping out a few words). There were also a handful of generations which provided incorrect paraphrasing (changed meaning of sentence extremely - see example). 
\begin{table*}[ht]
    \centering
    \begin{tabular}{p{0.37\linewidth}p{0.37\linewidth}p{0.26\linewidth}}
      \textbf{Original Text} & \textbf{GPT3.5 Generation} & \textbf{Category of Generation}  \\ \hline
      Though several attempts were made to found agricultural outposts with white settlers in the interior, these were largely failures. The oppressive tropical climate and hostile African neighbors made life difficult for settlers, many of whom lacked agricultural experience or expertise.&
Several attempts at establishing agricultural settlements with white settlers in the interior were attempted, yet were mostly unsuccessful. Settlers in the oppressive tropical climate and with hostile African neighbors found life to be a challenging experience, especially with many lacking any agricultural knowledge.&
Stylometric\\ \hline
Another issue is the added levels of coordination that occur when working with multiple aid groups to ensure healthcare service coverage. & An additional problem is the extra layers of organization that transpire when collaborating with numerous relief agencies to guarantee health care assistance. & Stylometric\\ \hline
The FBI databases contain tens of thousands of records from car-rental companies, hotels, and national department stores. & The databases of the Federal Bureau of Investigation include a plethora of files from car-rental establishments, hotels, and national chain stores. & Paraphrasing (missing content)\\
\hline
In Angola however, black Africans never made up less than $95\%$ of the population, so the demographic situation there (and in the other Portuguese colonies) was never the same as it was in Brazil. &In Angola, African-descended individuals always constituted in excess of $95\%$ of the populace, thus differentiating the demographic make-up of the Portuguese colonies from that of Brazil. & Incorrect Meaning
    \end{tabular}
\caption{Example of generations from GPT3.5 and the category of obfuscation method used.}
        \label{table: gpt3.5_examples}
\end{table*} 
\section{Time Consumption Analysis}\label{appx:time_consumption}
\begin{figure}[h!]
\begin{flushleft}
\includegraphics[width=.5\textwidth]{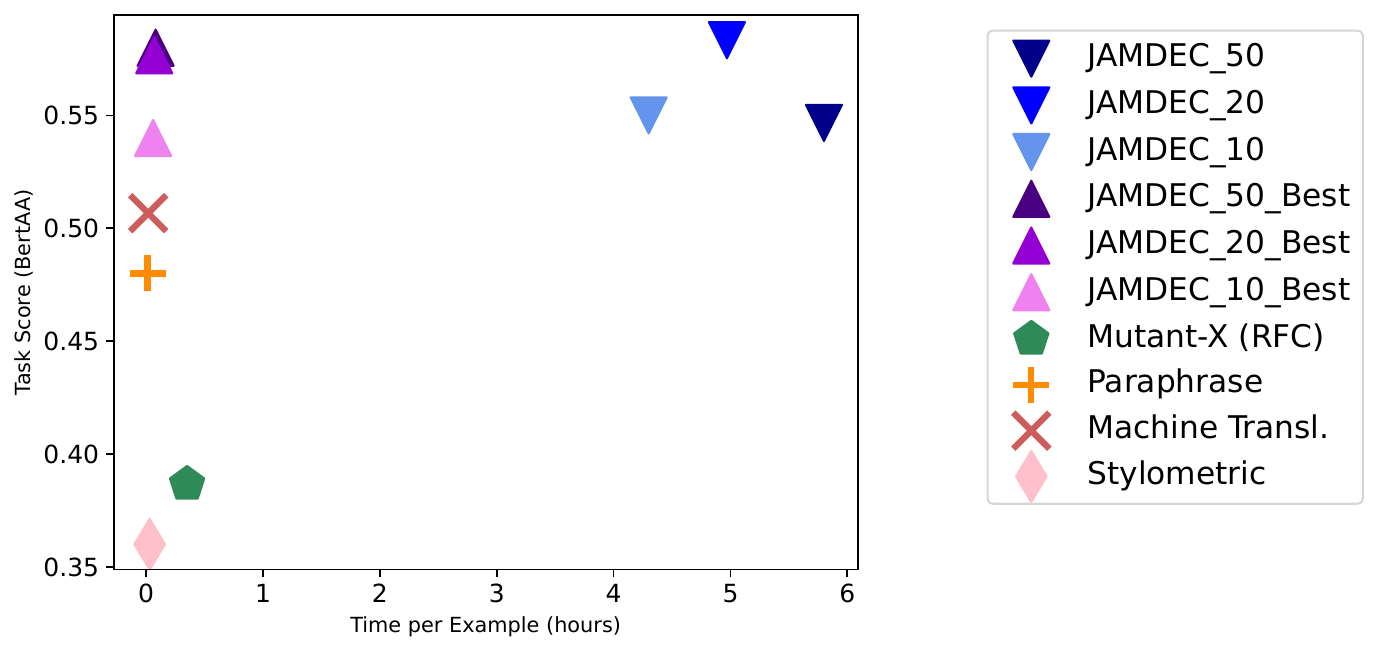}
\caption{Comparison of time consumption (hours) and performance (Task Score - BertAA). We compare \ourmethod (using all parameters of generations) and \ourmethod\_Best (using the best combination of generation parameters) to all other baseline methods.}
\label{fig:time_analysis}
\end{flushleft}
\end{figure}

We include a comparison of time consumption across the different obfuscation method. However, we recognize that there is a significant trade-off between time consumption and performance. Therefore, we provide, \cref{fig:time_analysis} which clearly illustrates this trade-off. 

In this analysis we showcase alter aspects of \ourmethod, beam width and generations parameters, which severely affect time consumption. First, we experiment with various beam width of 50, 20, and 10. We observe that when we reduce the beam size, the time consumption decreases significantly, yet the performance remains similar. Second, we experimented with using all parameter combinations versus using only the best parameter to generate candidates for filtering. Surprisingly, by using only the best parameter to generate a small candidate set which cuts the runtime by approximately five times, we achieve performance that's comparable to or even better than using all parameter combinations to produce a large candidate set.
Both ablations showcase the efficiency and effectiveness of \ourmethod. Additionally, when compared to other baselines, the best configuration of \ourmethod achieves significantly better performance with a comparable run-time. This further confirms the effectiveness and practicality of \ourmethod for real-world applications.

\section{Compare Similar Authorship Tasks}\label{appx:compare_author_tasks}

Here, we would like to further discuss the critical difference between seemingly similar language tasks: authorship obfuscation, paraphrasing and style transfer. Table \ref{tab:task_differ} provides a visual illustration of the differences in the tasks.

\paragraph{Paraphrasing} The main objectives of paraphrasing is to rephrase text to enhance clarity. Hence, paraphrasing can often lead to small edits that stay within the same authorship style, making it ineffective for concealing the author's identity. We further validate the incompetence of paraphrasing methods for authorship obfuscation empirically through both quantitative and qualitative analysis as shown in Table \ref{tab:amt_class_metric}, Figure \ref{fig:qual_examples} and Figure \ref{fig:qual_examples2}.

\paragraph{Style Transfer} Style transfer assumes a distinct target style
whereas authorship obfuscation assumes \emph{lack of} distinct style.
Specifically, while style transfer has a fixed target style as a priori, authorship obfuscation requires a dynamically changing output style depending on the particular input text to obfuscate. This makes it challenging to use style transfer techniques for authorship obfuscation, as it's hard to assume a specific target corpus representing the proper output style for obfuscation.
We further confirm the incompetence of style transfer methods empirically through quantitative and qualitative analysis as shown in Table \ref{tab:style_transfer} and Figure \ref{fig:qual_examples2}. In addition, using style transfer techniques for authorship obfuscation raises ethical concerns. The intention of authorship obfuscation is to safeguard the author's identity, avoiding the imitation or deceptive portrayal of an individual. Using style transfer to mimic another author could unintentionally blur the boundary between preserving anonymity and indulging in deceitful behavior.

\begin{table*}[t]
    \centering
    \begin{tabular}{lcccc}
    \toprule
   \textbf{Task} & \textbf{Preserve All Content} & \textbf{Preserve Tone}& \textbf{Change in Style}& \textbf{Target Style}\\
  
    \hline
    Authorship Obf. & \cmark & \cmark & \cmark & \xmark\\
    Paraphrase & \xmark & \cmark & \xmark & \xmark\\
    Style Transfer& \cmark & \cmark & \cmark & \cmark\\
    \end{tabular}
    \caption{Comparison between the task of authorship obfuscation, paraphrasing and style transfer.}
    \label{tab:task_differ}
\end{table*}  %
\section{Experimental Details}\label{appx:exp_details}
In this section we provide full details of the experimentation used in this paper. We start with the dataset in Appendix \ref{appx:exp_data}, method implementations and hyperparameter choices for each method in Appendix \ref{appx:exp_hyperparam}, and evaluation methodology in Appendix \ref{appx:exp_eval}. 

\subsection{Data}\label{appx:exp_data}

\myparagraph{AMT- Formal Articles}
The dataset, the Extended-Brennan-Greenstadt \cite{amt_dataset}, contains collections of short ($\sim 500$-words) scholarly text that were gathered from Amazon Mechanical Turk (AMT). These articles were collected using very strict guidelines which required the writing to be clear (free of citations, urls, headings, etc.), true to the author's writing style, relevant to the topic, and the correct length. These qualities were then reviewed by the researchers after submission for quality assurance. More information about the data collection can be reviewed in \citet{amt_dataset}. We used the same three test sets as \citet{mahmood2019mutantx}, which were a collection of 3, 5, and 10 authors with 27, 30, and 49 texts respectively (\amt{3}, \amt{5}, \amt{10}). Each author wrote about the same topic throughout the different text. Examples of the author's topics included identity theft, and Portuguese slavery in Africa. An example of a passage can be seen in \cref{tab:dataset_examples}.

\myparagraph{BLOG- Informal Articles}
The second dataset, the Blog Authorship \cite{blog_dataset}, contains a collection of blog entries that were posted to blog.com in 2004. The original dataset contains over 680k post from 19k individual authors, with an average of 7,250 words per author. Each author tends to write about similar topics and styles, ranging from dairy style entries to fan-fiction. Similar to the test sets used by \citet{mahmood2019mutantx}, we created two datasets with a collection of 5, and 10 authors with 72, and 150 texts respectively (BLOG-5, BLOG-10). An example of a passage can be seen in \cref{tab:dataset_examples}. 

\begin{table*}[h!]
\centering
\begin{tabular}{p{1cm}p{15cm}}
 \hline
 \textbf{Dataset} & \textbf{Text Example }\\ [0.5ex] 
 \hline
 AMT &  In the 1990s Zaire served as the main supporter of UNITA, as South African and American support for the organization dwindled. In 1997 a coup supported in part by the Angolan government overthrew Mobutu, and Zaire was renamed the Democratic Republic of the Congo. Without the aggressive Mobutu regime as a neighborhood, the situation in Angola stabilized and the MPLA was finally able to crack down on internal dissent without being troubled with foreign intervention, ending the civil war a few years later in 2002. Like most other Third World conflicts of the twentieth century, the wars in Angola were heavily affected by the Cold War. In addition to the competition between the US and the USSR, several other factors motivated the involvement of international powers: the Sino-Soviet split, Third World solidarity against Western exploitation and imperialism, and in the case of the US, Angola's large oil reserves. The USSR was involved with the MPLA from its foundation in the late-1950s. Starting in 1958, MPLA founding member Mario de Andrade would travel to Moscow on a regular basis for various conferences and meetings. During these visits the MPLA developed a relationship with the Soviets, securing funding and in 1961 the explicit support of Soviet Premier Nikita Khrushchev, who stated that ‚"the patriots of Angola can be sure that the sympathies of the peoples of the great Soviet Union are fully on their side." Many MPLA leaders would go on to be educated in Moscow. The USSR chose to support the MPLA over rival movements in Angola for a number of reasons. As a left-leaning Marxist movement that explicitly condemned the imperial powers, the MPLA followed the same basic ideological principles as the USSR. The UPA/FNLA was more ambiguous on this issue, receiving support from the US and sometimes practicing anti-communist rhetoric. The MPLA was also not as focused on regional or ethnic issues, as the predominately Bakongo UPA based in northern Angola was. The USSR also practiced the policy of recognizing and supporting only one rebel movement within a conflict, a policy not shared by all of its peers. Early Soviet support of the MPLA included food and clothing as well as weapons and increased progressively during the course of the war from goods valued at \$25,000 in 1961 to \$220,000 in 1973. Large scale Soviet assistance did not come until 1975 though. In this year another foreign power would join the equation, with Cuba‚ shipment of two shiploads of T-55 tanks and 500 military advisories. Though the Cubans and Soviets would work together closely in Angola, early actions were not coordinated as is widely assumed. Cuba was not simply a Soviet proxy but rather had its own agenda for being in Angola. As a Third World country with a colonial past and communist government, Cuba wanted to sustain the global conflict against the West and imperialism through spreading Marxist-Leninist revolution. \\ 
 \midrule
 BLOG & 7:05 a.m.  Wednesday.  Feeling pretty good today.  My last couple hours of sleep were choppy, but I went to bed so early I’m sure I got at least eight hours.  Took half an actifed to counter the red wine, and I didn’t drink enough water to counteract them both.  Other than that, feeling good, and I’m pleased with the amount I drank for Drinking Night.  My new plan is to buy only red wine, and buy only enough for the one drinking night.  If I don’t have it around the house, I won’t drink it.  Because I am far too lazy and too self-conscious to go buy it.  Therefore, this way I am not relying on willpower, I’m setting up an environment where I can’t drink.  I’m having a glass of water right now, with my coffee.  I don’t usually start until after breakfast, but I feel quite dehydrated.  I’m adjusting my estimates for the coffee with Benefiber, because I’m not putting an entire tablespoon in.  Maybe two-thirds that.  Note: remember to buy an exercise ball to sit on while at the computer.  5:00 p.m.  Had a nice little lunch with Daisy.  Ate a veggie wrap and some fries, which I hope I am estimating reasonably.  It was a decent meal, but not entirely filling, so I had a little chicken when I got home.  Now I am finishing up my work emailing before vacation, trying to do my timesheet, etc.  My hip is still bothering me.  I’m not happy about that, because it hurts when I walk, and I want to do a lot of walking on vacation.  I think the bellydancing may have caused the strain, and then the gliding is exacerbating it.  So perhaps it’s a good thing that I’ll be away from the glider for a couple weeks.  I can walk and swim for exercise, and perhaps that will work out the problem, whatever it is. \\
 \hline
\end{tabular}
\caption{Examples of text from both datasets used in the experimentation section; AMT and BLOG.}
\label{tab:dataset_examples}
\end{table*} 
\subsection{Method Implementation}\label{appx:exp_hyperparam}
The method implementation and hyperparameters for each method used in our experimentation are detailed below. 
\subsubsection{Baselines} \label{appx:baselines}\xspace 
\myparagraph{Stylometric Obfuscation} We employ the Stylometric Obfuscation method proposed by Karadzhov et al. \cite{stylo_method} in the PAN-2016 Author Masking Shared Task competition \cite{PAN2016}. This method calculates metrics for 12 features that are indicative of style, then modifies the text, so these metrics align with an "average" value. The "averages" were calculated using a combination of training sets including the PAN-2016 Author Obfuscation task \cite{PAN2016} and public domain books from Project Gutenberg \cite{Project_Gutenberg} Examples of the metrics this method uses include the average number of words per sentence, word frequency, and the use of uppercase letters. Changes employed include actions such as sentence splitting and merging, substitution of words with synonyms, and alterations in spelling. For a full list of metrics and proposed changes, see the \cite{stylo_method}. To further enhance the obfuscation process, the method introduces "noise" by modifying words that differ between English and British English and introducing additional functional words. We make no changes to the hyperparameters used in the original method. 

\myparagraph{Mutant-X} Mutant-X \cite{mahmood2019mutantx} is a genetic algorithm guided by an internal authorship classifier, which is trained using a deep learning architecture. This method requires a separate authorship corpus to train the internal classifier. The approach follows an iterative process wherein the text undergoes "mutation" by randomly replacing words based on their semantic similarity. Additionally, at random iterations, it incorporates a "crossover" effect that involves cutting two parent texts at a random position and combining them to create two new child texts. 

In order to train the internal classifiers, we adopt the same features and architectures (Random Forest Classifier, Support Vector Classifier and Ensemble Classifier), as suggested in the subsequent work by \citet{avengers}. The features come from the Writeprints-Static \cite{amt_dataset} which is a set of $500+$ lexical and syntactic features such as total words, percentage of digits, and special character occurrences taken from the full set Writeprint \cite{writeprint}. For the Random Forest (RFC) we used 50 trees in the forest and all features, and for the Support Vector Classifier (SVC) we use a linear kernel with all the features. However, as was explored in \citet{avengers}, a single classifier may not be the best indication of authorship classification since it could be relying on artifacts to make its decision. Therefore, they introduced the idea of using an "ensemble" of classifiers (ENS) which are trained on different collections of features and vote on the final classification. We used an ensemble of $10$ SVC classifiers which were trained on 30 randomly selected features. A "soft" voting system (averaging probabilities instead of classes)  was used to choose the final obfuscation. However, we only show results with an internal classifier of RFC or ENS for the internal classifiers. The sample size,  train accuracy, and test accuracy of these models can be reviewed in \cref{tab:classifier_acc}.

\begin{table*}[t]\centering
    \resizebox{1\textwidth}{!}{
    \begin{tabular}{ lccccccccc }\toprule
        
       \multicolumn{1}{c}{\textbf{Dataset}}&\multicolumn{1}{c}{\textbf{Train Sample Size}}& \multicolumn{1}{c}{\textbf{Test Sample Size}}&\multicolumn{2}{c}{\textbf{ENS } }& \multicolumn{2}{c}{\textbf{RFC}} & \multicolumn{2}{c}{\textbf{BertAA}}\\
        
        \cmidrule(lr){4-5}\cmidrule(lr){6-7}\cmidrule(lr){8-9}
       &&&\textbf{Train Acc. } & \textbf{Test Acc.}  &\textbf{Train Acc. } & \textbf{Test Acc.} &\textbf{Train Acc. } & \textbf{Test Acc.} \\  
        \midrule
\amt{3}&36 & 27& 1.0 & 0.93 & 1.0 & 0.93 & 1.0 & 0.93\\
        \midrule
\amt{5} &60& 30&  1.0 & 0.93 & 1.0 & 0.87 & 1.0 & 0.87\\
        \midrule
\amt{10}& 120 &49 &  1.0 & 0.82 & 1.0 & 0.69 & 1.0  & 0.57 \\
\midrule
BLOG-5 &400& 100&  1.0 & 0.93 & 1.0 & 0.91 & 1.0 & 0.98\\
        \midrule
BLOG-10 &  800& 150& 0.96 & 0.84 & 1.0 & 0.83 & 1.0 & 0.95\\

        \bottomrule
    \end{tabular}
    }
    \caption{Train and test accuracy for the three classifiers used in the experimentation (ENS, RFC, and BertAA) for each dataset (\amt{3}, \amt{5}, \amt{10}, \blog{5}, \blog{10}). We also display the sample size for the training and test set for each dataset.}
    \label{tab:classifier_acc}
\end{table*}

\myparagraph{Paraphrasing} For the paraphrasing baseline, we employ a state-of-the-art paraphrasing model, PEGASUS Paraphrase \cite{pegasus, paraphrase_model} a PEGASUS model fine-tuned on a self-supervised task for paraphrasing. 

\myparagraph{Machine Translation} Inspired by the work of Keswani et al. \cite{Keswani2016AuthorMT}, we implemented a similar approach using machine translation from English to German, then to French, and finally back to English. Keswani et al. emphasized the importance of using a machine translation model that does not rely on English as an intermediate step. This means that when translating from German to French, the model should go directly from German to French, without translating via English. In their paper, they did not provide the code for this method, so we created our own implementation using the M2M100 translation model \cite{m2m100} with 418M parameters.

\textbf{GPT3.5} We include a comparison with zero-shot prompting using GPT-3 (text-davinci-003, 175B) 3 \cite{gpt3} which has $\sim 175$B parameters. Our comparison involved prompting at both the sentence-level, where each sentence was obfuscated individually, and the paragraph level, where the entire text was obfuscated as a whole. We prompted GPT-3 to generate two obfuscations for each sentence/paragraph. Subsequently, for the sentence-level obfuscation, we randomly combined one generation from the two produced for each sentence to create a single obfuscated paragraph. The evaluations presented here represent the average performance across these two generations. However, due to financial constraints, we limited our GPT-3 obfuscation generation to AMT-3.

Below are the exact prompts used to generate obfuscated text at the sentence and paragraph level.

\textit{Sentence-level:} 

"Provide two re-writes of the following sentence so that the author's style is obfuscated. 

Original Sentence: \{original text\}"

\textit{Paragraph-level:} 

"Provide two re-writes of the following paragraph so that the author's style is obfuscated. 

Original Paragraph: \{original text\}"

\subsubsection{\ourmethod}As described, \ourmethod has three distinct stages (keyword extraction, over-generation, and filtering). We also include a pre-processing step which prepares the raw data for obfuscation. We outline the hyperparameter values used in each section below.

\myparagraph{Data Pre-Processing}
We pre-process the raw text before obfuscating. First, we divide each text into paragraphs. We go through each sentence in each paragraph and add it to a list $y_{\text{orig}}$. We then group all sentences in that same paragraph that appear previously and store it in a new list $x_\text{l}$. This results in a list of original sentences $y_{\text{orig}}$ and left contexts $x_\text{l}$. If the sentences are the first in the paragraph, we use the previous's paragraphs last sentence as the left context. For the first sentences of the text, we use itself as the left context. Lastly, if a sentence has less then $3$ words we did not change it. 

\myparagraph{Keyword Extraction}
We use three kinds of keyword extraction; KeyBERT, Likelihood-T5 and Likelihood-GPT2 as described in \cref{sec:expt}. For KeyBERT we used unigrams and returned $n/2$ keywords, where $n$ was the length of the original sentence. For Likelihood-T5, we used a T5-base \cite{t5} and for Likelihood-GPT2 we used a GPT2-XL (1.5B) \cite{gpt2}. For both Likelihood-T5 and Likelihood-GPT2, we used a likelihood threshold of 0.5, meaning any original word whose next token probability was below 0.5 was kept as a keyword. 

To further support creative and diverse generation, we include disjoint constraints which allow for one of a list of constraints to be met. Using disjoint constraints, we add both "like" words (same root word with different tenses) and "similar" words (synonyms) of the keywords. To do this, we start by creating a static dictionary of word embedding. For our experimentation, we used a list of 20K most common English words  \cite{mostcommonwords} and convert each word into the tokens using T5-base pretrained model \cite{t5}. For more details on this static dictionary see \cref{appx:stylometric-based-obfuscator}. Then, to find the top "similar" words, we used the cosine similarity between the original keyword and each word in the static dictionary and choose the top $4$ with the highest score. To find the top "like" words, we used the Spacy package \cite{spacy2} in Python to find the first $4$ words in the static dictionary with the same word lemma as the original keyword. For our experimentation, we used three versions of the keywords as constraints. We used the original keywords, the original keywords with the "like" words, and the original keywords with the "like" and "similar" words.

\myparagraph{Generation}
For our experimentation, we used Neurologic Constrained Beam Search \cite{neurologic} and Diverse Beam Search \cite{vijayakumar2018diverse}. The base model was GPT2-XL (1.5B) For most of the experimentation (except for the ablation study in \cref{appx:lightweight_exp}), we used a beam width of $50$ and a matching number of return sequences. The maximum length of the generation was set to twice the largest input length in a batch. The batches were grouped by input length, to keep like max lengths. We also set the no repeat length to $3$-grams. For decoding within the beam search, we ran each combination twice, once with sampling decoding and another with greedy decoding. We used a likelihood pruning factor of $0.4$ and a constraint pruning factor of $0.6$. For the constraints, we used both ordered constraints (the constraint must be met in a specific order) and unordered constraints. Lastly, we employed early stopping, which will stop a beam search early if candidates are not better than the current candidates. When diversity was employed, we used a diversity penalty of $5,000$. Hyperparameters were selected based on experimentation on Reuter 50-50 \cite{misc_reuter_50_50_217}, which is a sub-sample of newswire articles produced by Reuters in 1996 - 1997 which have at least one subtopic of class corporate/industrial. This is a common baseline used for authorship verification \cite{Qian2017DeepLB}.

In summary, we ran generations for each sentence using the following combinations of methods:
\begin{itemize}
    \item \textit{Decoding Method}: Sampling, Greedy
    \item \textit{Type of Constraints}: Original, Original + Like, Original + Like + Similar
    \item \textit{Ordered Constraint}: True, False
    \item \textit{Diversity in Pre-Processing}: True, False

\end{itemize}
\myparagraph{Filtering}
For our experimentation, we ran two different filtering techniques. Each method starts with a base NLI and CoLA threshold. Due to the lack of an evaluation set, all hyperparameters were selected using a grid search on the smallest dataset of each kind (AMT-3 and BLOG-5). In some cases, we find that none of the generated candidates passes both the NLI and CoLA filter. To process such cases, we consider two variants of our method: (1) \ourmethod, where we simply output the original sentence as output, and (2) \ourmethod + Stylo, where we run a basic stylometric-based obfuscator on the original sentence and then use a second CoLA threshold for this altered sentence. The basic stylometric-based obfuscator is explained in detail below in \cref{appx:stylometric-based-obfuscator}. If the altered sentence does not pass the filer than the original sentence is used. A full list of hyperparameters for each method can be viwed in \cref{tab:hyperparam}. We also provide the average percentage of sentences that passed the basic NLI/CoLA thresholds and the second CoLA threshold that is used in \ourmethod + Stylo in \cref{tab:generation_pass_thresholds}. 

\begin{table}[t]
    \centering
    \resizebox{.5\textwidth}{!}{
    \begin{tabular}{llcc}
    \toprule
   \textbf{Dataset} & \textbf{Hyperparameter} & \textbf{\ourmethod} & \textbf{\ourmethod + Stylo}\\
    \hline
    AMT&Base NLI Thresholds& 0.30 & 0.40\\
    &Base CoLA Threshold & 0.30 & 0.40\\
    &Second CoLA Threshold & - & 0.70\\
    \midrule
    BLOG&Base NLI Thresholds& 0.10 & 0.10\\
    &Base CoLA Threshold & 0.10 & 0.10\\
    &Second CoLA Threshold & - & 0.70\\
   \bottomrule
    \end{tabular}
    }
    \caption{Hyperparameters for the filtering stage of the experiments using \ourmethod with and without the stylometry decoding (+ Stylo); AMT and BLOG datasets}.
    \label{tab:hyperparam}
\end{table} %
\begin{table*}[t]
    \centering
    \begin{tabular}{llll}
    \toprule
   \textbf{Dataset} & \textbf{\ourmethod} & \textbf{\ourmethod + Stylo}\\
    \hline
    AMT-3& Pass Base Thresholds& 0.52 & 0.63 \\
    &Pass Second CoLA Threshold & - & 0.15\\
    & Original Sent. Used & 0.48 & 0.22\\
    \midrule
    AMT-5&Pass Base NLI Threshold& 0.52 & 0.64\\
    &Base Pass CoLA Threshold & - & 0.16\\
        & Original Sent. Used &0.48&0.20\\

        \midrule
    AMT-10&Pass Base NLI Threshold& 0.53 & 0.60\\
    &Base Pass CoLA Threshold & - & 0.13\\
        & Original Sent. Used &0.47&0.27\\

        \midrule
    BLOG-5&Pass Base NLI Threshold& 0.57 & 0.64\\
    &Base Pass CoLA Threshold & - & 0.07\\
        & Original Sent. Used &0.43&0.29\\

        \midrule
    BLOG-10&Pass Base NLI Threshold& 0.60 & 0.67\\
    &Base Pass CoLA Threshold & - & 0.06\\
        & Original Sent. Used &0.4&0.27\\

   \bottomrule
    \end{tabular}
    \caption{Breakdown of average number of sentences that pass both the base thresholds (NLI and CoLA), the second CoLA threshold (only used for \ourmethod + Stylo), and the average original sentences used for each dataset.}
    
    \label{tab:generation_pass_thresholds}

\end{table*} 
\subsubsection{Our Stylometric-Based Obfuscator}\label{appx:stylometric-based-obfuscator}
\myparagraph{Set-Up}
We consider the original prompt (sentence) $x$ which is composed of words $x_1, ..., x_n$. Before decoding, we "freeze`` all tokens that correspond to function words. Function words are grammatical words that serve as connectors or structure indicators in a sentence, rather than conveying lexical meaning. Therefore, we only consider changing context words such as nouns, adjective, and verbs. A difficult aspect of a word-changing method is choosing which words are truly equivalent to the original word. For our method, we consider new words as replacements based on the following:
\begin{enumerate}
    \item Similarity to the original word $S_t$
    \item Grammatical correctness of new sentence $G_t$
\end{enumerate}

\begin{figure}[h]
\centering
\includegraphics[width=0.5\textwidth]{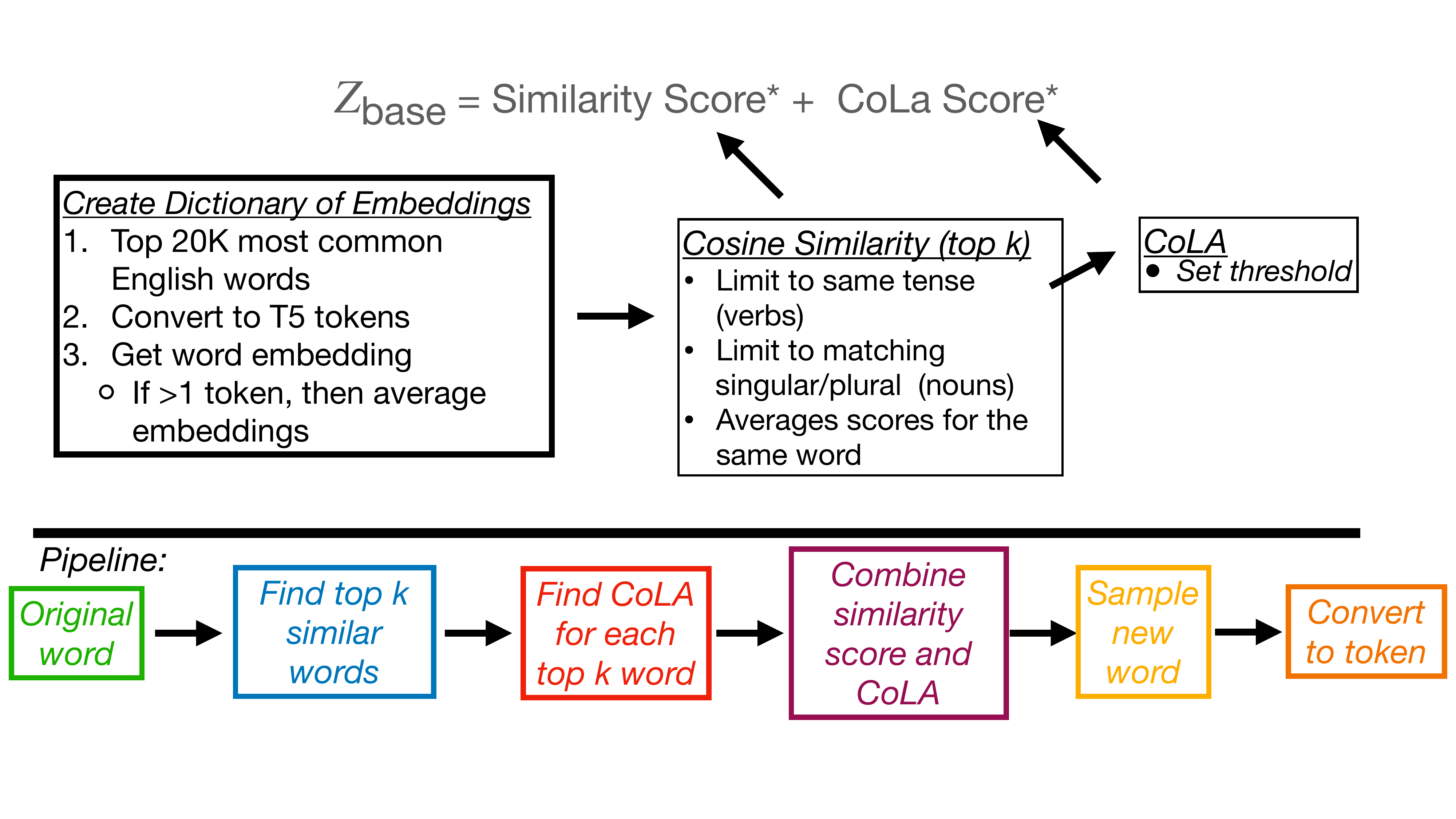}
\caption{A visual representation of the pipeline used for the stylometric-based obfuscation method used in \ourmethod+Stylo.}
\label{fig:thesaurus_pipeline}
\end{figure}

Using these two metrics, we created a 3-step method for identifying and changing certain words of a sentence. The pipeline can be viewed in Figure \ref{fig:thesaurus_pipeline} and is described in detail below. 

\paragraph{Step1: Word Embeddings Dictionary}We start by creating a new static dictionary of word embedding, depending on the base model. For our experimentation, we use a list of 20K most common English words  \cite{mostcommonwords} and convert each word into tokens using T5-base (220M) pretrained model \cite{t5}. Then, using these matched tokens, we extracted their corresponding word embedding vectors (weights in the last attention layer). If a word matched to multiple T5 tokens, then we averaged their corresponding word embedding vectors. This resulted in a static word embedding dictionary D of vectors $d_1, ..., d_{20K}$, where $d_i \in \mathbb{R}^{|V|}$, where, $V$ is the length of the T5 vocabulary. 
\paragraph{Step 2: Similar Words} Next, we find the top $k$ similar words from $D$ to the original word $x_t$ using cosine similarity of the word embeddings. We only consider verbs of the same tense and nouns that match the singular or plural nature of the original token $x_t$. Let $W$ be the set of words $w_1, ..., w_k$ with the highest similarity scores $s_i$. With this set $R$ of top-$k$ similarity scores, $s_1,...,s_k$,  we create the following similar score distribution $S_t$ for original word $x_t$
\begin{align}
    S_t=
    \begin{cases}
       \frac{s_i - \min(R)}{\max(R) - \min(R)}  & \text{if } w_i \in W\\
        0 & \text{otherwise}.
    \end{cases}
\end{align}
\paragraph{Step 3: Grammar Scores} Using the top $k$ similar words $w_i, ..., w_k$ from the previous step, we find each grammar score $g_i$ using a Roberta base model \cite{roberta} finetuned on the Corpus of Linguistic Acceptability (CoLA) \cite{cola, textattack}, a large corpus which contains 10.5K sentences annotated for grammar acceptability by their original authors. We do this by using the \textit{generated text} $x_1,...,x_{t-1}$ before $x_t$, and using the \textit{original text} $x_{t+1},...,x_n$ after the generated text. For example, if the original text was "I went to a big lake", and we have generated "I walked to a" and are currently trying to find the grammar score for "huge", we would use "I walked to a [huge] lake" as input to the CoLa model. We use the probability of the input being grammatically acceptable as $g_i$. We do this for each similar word, resulting in a set $Q$ of grammar scores $g_1, ..., g_k$. Lastly, we impose a lower threshold $\delta$, which we set, so the grammar scores are guaranteed to be high. This can be tuned for specific tasks.  Similar to the similarity scores, we construct a grammar score distribution $G_t$ for the original word $x_t$ as 
\begin{align}
    S_t=
    \begin{cases}
       \frac{g_i - \min(Q)}{\max(Q) - \min(Q)}  & \text{if } w_i \in W, g_i > \delta\\
        0 & \text{otherwise}.
    \end{cases}
\end{align}

\paragraph{Step 4: Word Selection} Lastly, we combine the similar score distribution $S_t$ and grammar score distribution $G_t$ using the following equation,
\begin{align}
    F_t = \alpha S_t + \beta G_t
\end{align}
where $\alpha$ and $\beta$ are hyperparameters controlling the importance of similarity or grammatical acceptability. We use sampling from the final distribution, $F_t$ to generate the word replacement. However, we note that the original word is included in the top $k$ similarity and therefore could result in the final generation. This method is repeated for each context word from the original text. An example of this method on text from the Reuter 50-50 dataset \cite{reuter5050} can be found in \cref{table: thesaurus_method}. 

\newlength\mylength
\setlength\mylength{\dimexpr.5\columnwidth-2\tabcolsep-0.5\arrayrulewidth\relax}

\begin{table}[t]
    \begin{tabular}{p{\mylength}|p{\mylength}}
        \hline
        \textbf{Original Text} & \textbf{Obfuscated Text}    \\
        \hline
 The site does not include the countries' actual data -- that 
may come later -- but it lists contacts for obtaining the information. & The site does not \textbf{contain the states' real files} -- that 
\textbf{might} come later -- but it \textbf{includes} contacts for obtaining the information.\\
        \hline
The International Monetary Fund open a site on the Internet Thursday providing 
information about the types of economic
 data available in 18 member countries. &
 The International Monetary Fund \textbf{started} a \textbf{page} on the internet Thursday \textbf{delivering advice} about the types of economic \textbf{records offered in 18 membership regions}. \\
 \hline
 Senator Bob Kerrey is preparing legislation in an attempt to break the deadlock over computer encryption export 
policy, people familiar with the Senator\'s plans said. &
Senator Bob Kerrey is preparing regulation in an \textbf{effort to crack} the deadlock over \textbf{internet} encryption \textbf{importation} 
policy, people \textbf{acquainted} with the Senator's plans said.\\
\hline
    \end{tabular}
        \caption{Example of sentences obfuscated using our basic stylometric-based obfuscator. On the left is the original text and on the right is the obfuscated text. The changes are show in \textbf{bold}.}
        \label{table: thesaurus_method}
\end{table}

\subsection{Evaluation Methodology and Other Details}\label{appx:exp_eval}
\myparagraph{Automatic Evaluation}\label{appx:automatic_evaluation}
We used five automatic evaluations; Drop Rate (ENS and BertAA) \cite{mahmood2019mutantx, fabien-etal-2020-bertaa}, METEOR \cite{meteor}, NLI \cite{liu-etal-2022-wanli}, and CoLA \cite{cola}. The Drop rate is the average decrease in number of obfuscated text which a classifier identified as the non-original author compared to the original text. Two classification models were used to calculate the drop rate, an ENS and BertAA model. The training of ENS model is described in \cref{appx:baselines} under "Mutant-X" \cite{mahmood2019mutantx}. The training for BertAA is described in \cite{fabien-etal-2020-bertaa}. METEOR (Metric for Evaluation of Translation with Explicit ORdering) \cite{meteor} is a common baseline used in machine translation. It is calculated using the harmonic mean of precision and recall using unigram matching that ranges from 0 (no overlap) to 1 (exact overlap). Because it relies on exact token matching, it is unideal for measuring paraphrases of text that could have drastically different tokens but the same meaning. We include the reporting of this metric since it is heavily reported in the literature. However, we rather rely on another metric, NLI (Natural Language Inference) as an indicator of content preservation. NLI is a task with aims to predict if two text are "entailed", in other words if one text is true then the other logically follows. We used WANLI model \cite{liu-etal-2022-wanli} as our NLI model and report the average highest NLI scores for each sentence. Meaning, we take each sentence in the obfuscated text and calculate the probability of entailment, according to the WANLI model, with each sentence in the original. We then choose the highest entailment value. What is reported is the average of these maximum values for all text. Lastly, we use a CoLA (Corpus for Linguistic Acceptability) \cite{cola} model as a measure of grammatical correctness. Given a text, the model reports a probability of grammatical acceptance (ranging from 0 to 1), we use the average of these as the CoLA score. 

\myparagraph{Inter-rater Agreement} We decided to use two different classifier models (ENS and BertAA) to calculate the drop rate. Since these models use different architecture and different sets of features, we wanted to report the inter-rater agreement between them. We use Cohen's kappa coefficient, which measure the inter-rater reliability using a scale between $[0.1]$, where $0$ is completed disagreement and $1$ is complete agreement. This is thought to be a more robust measure because it takes the probability of agreement by chance into consideration. See \cref{tab:inter_rater_agreement} for the results. 

\begin{table*}[t]\centering
    \resizebox{\textwidth}{!}{
    \begin{tabular}{ llccccccccc }\toprule
        & \multicolumn{1}{c}{\textbf{Method}} & \multicolumn{2}{c}{Mutant-X} & \multicolumn{2}{c}{GPT3} & \multicolumn{1}{c}{Paraph} & \multicolumn{1}{c}{Machine Transl.}& \multicolumn{1}{c}{Stylometric}& \multicolumn{2}{c}{\method}\\
        
        \cmidrule(lr){3-4}\cmidrule(lr){5-6}\cmidrule(lr){7-7}\cmidrule(lr){8-8}\cmidrule(lr){9-9}\cmidrule(lr){10-11}
        
        \multicolumn{1}{c}{\textbf{Dataset}}&\multicolumn{1}{c}{\textbf{Classifier}} & \textit{ENS}  & \textit{RFC}   & \textit{Sentence} & \textit{Paragraph} & & & \textit{W/O Stylo} & \textit{W/ Stylo}  \\
        
        \midrule
        &ENS-RFC & 0.19 & 0.27 & 0.72 & 0.59 & 0.83& 0.82 & 0.77 & 0.66 & 0.67 \\
        AMT-3 &ENS-BertAA &0.83 &0.39&-&-&0.89& 0.65& 0.58 &0.77  &0.77  \\
        &BertAA-RFC & 0.30&0.72	&- &- &0.83  &0.65  &0.78  &0.89&0.89 \\
                \midrule
        &ENS-RFC & 026 & 0.33 & - & - & 0.57&0.60 & 0.54 & 0.64 & 0.69 \\
        AMT-5 &ENS-BertAA &0.09 &0.29	&- &- &0.54  &0.56  &0.53  &0.47&0.43 \\
        &BertAA-RFC & 0.44&	0.11&- &- &0.63  &0.47  &0.31  &0.50&0.54 \\
                        \midrule
        &ENS-RFC & .03 & 0.21 & - & - & 0.45& 0.39 & 0.57 & 0.39 & 0.35 \\
        AMT-10 &ENS-BertAA & 0.10 &	0.38& -&- &.56 & 0.34  &0.48  & 0.29 &0.36 \\
        &BertAA-RFC & 0.43&	0.11&-&-& 0.52& 0.34& 0.38 & 0.37 &  0.35 \\
        \bottomrule
    \end{tabular}
    }
    \caption{Inter-rater reliability score (Cohen kappa coefficient) between each classifier (RFC, ENS, and BertAA) used for the AMT dataset. }
    \label{tab:inter_rater_agreement}
\end{table*} 

\subsection{Human Evaluation}\label{appx:human_evaluation} 
\begin{figure*}
\begin{subfigure}[b]{0.5\textwidth}
    \includegraphics[scale=.48]{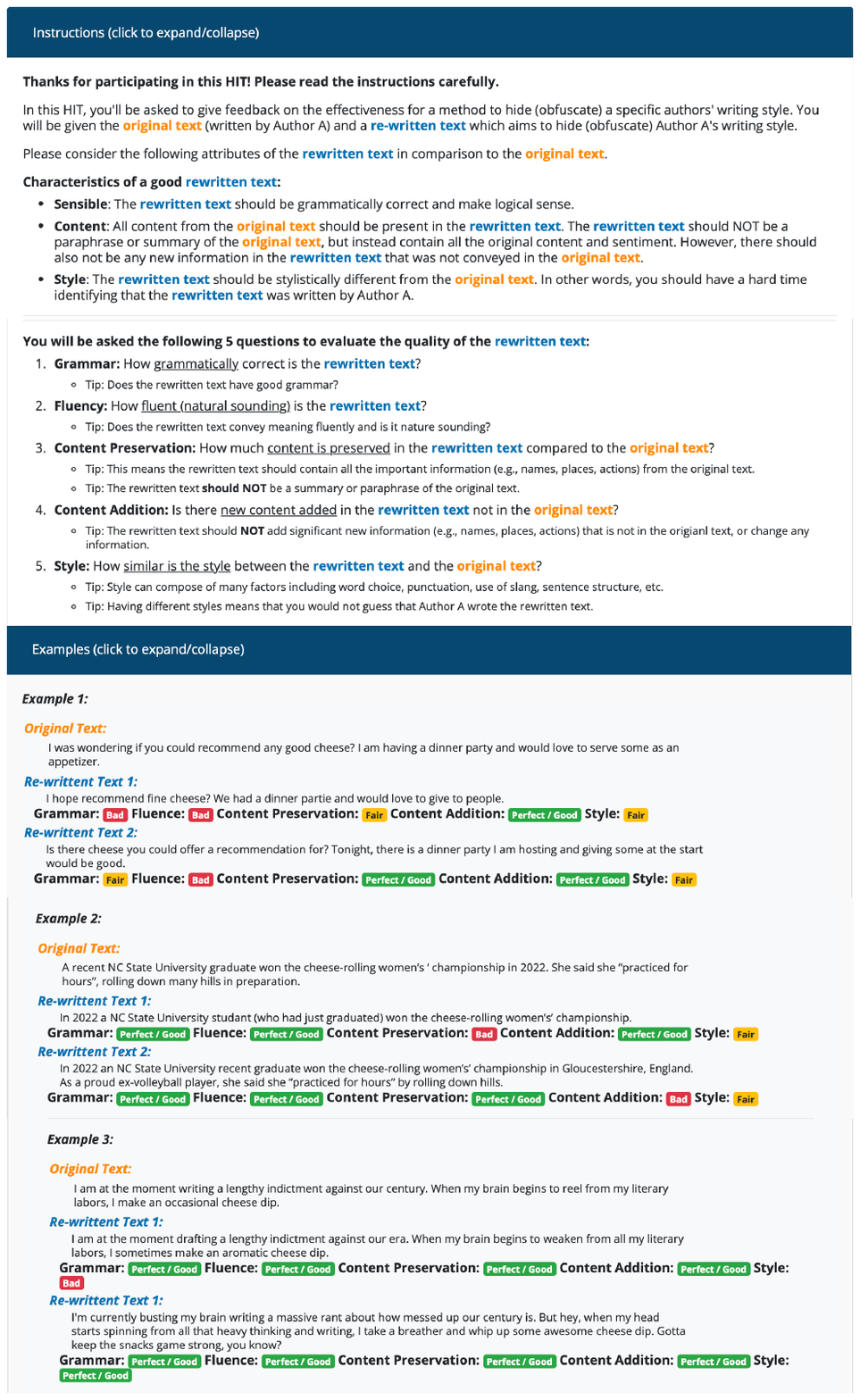}
    \caption{Instructions}
\end{subfigure}
\begin{subfigure}[b]{0.5\textwidth}
    \includegraphics[scale=.48]{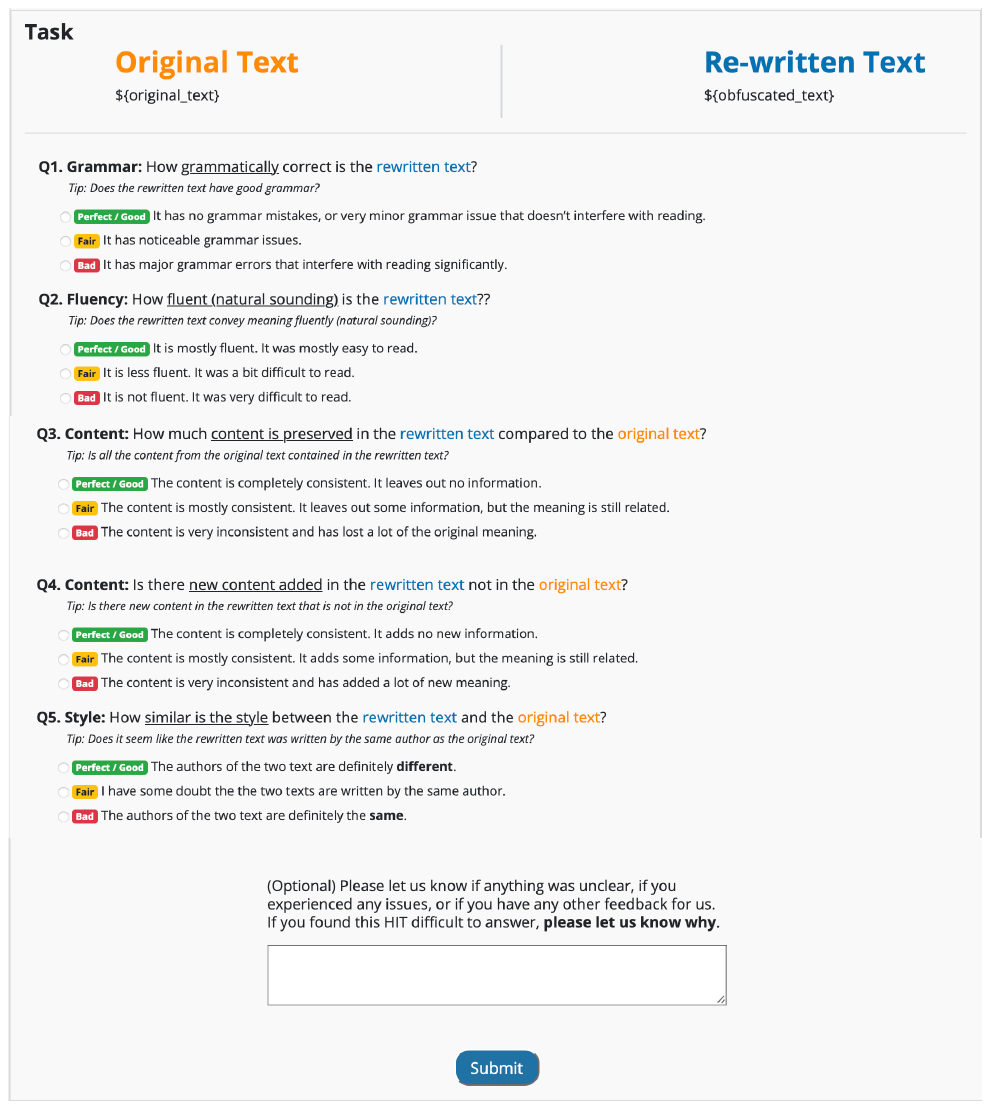}
    \caption{Task}
\end{subfigure}
\caption{Instructions and task for the human evaluation done through Amazon Mechanical Turk.}
\label{fig:amt_instrutions}
\end{figure*}

All human evaluations were conducted on Amazon Mechanical Turk (AMT) \cite{amt}. The data for the human evaluations were randomly selected from the passages in AMT-3. Each passage was separated into shorter sections ranging from one to four sentences. Then $n=32, 35,$ and $35$ of these shorter sections were selected from author "H", "PP", and "QQ" texts respectively (Author "H" has fewer passages overall than "PP" or "QQ" and therefore had slightly less short texts chosen for the human evaluation) for a total of $102$ passages. The corresponding obfuscated text was then matched for the following methods; Mutant-X (ENS), Machine Translation, Stylometric, GPT3.5 (Sentence), \ourmethod, and \ourmethod + Stylo. For each passage, the AMT worker was shown the original and obfuscate passage side by side and ask the following five questions. 
\begin{enumerate}
    \item Grammar: How grammatically correct is the rewritten text?
    \item Fluency: How fluent (natural sounding) is the rewritten text?
    \item Content: How much content is preserved in the rewritten text compared to the original text?
    \item Content: Is there new content added in the rewritten text not in the original text?
    \item Style: How similar is the style between the rewritten text and the original text?
\end{enumerate}
Each question was answered on a 3-point likert scale (Perfect/Good, Fair, and Bad). Detailed instructions and examples were provided, see \cref{fig:amt_instrutions}. We compensate workers with the hourly wage of $15$. We used a few credential checks for our Mechanical Turk workers. First, their HIT Approval Rate for all Request had to be greater than $97\%$ and they had to be pre-approved based on work they had done in other unrelated tasks from our lab. Due to financial constraints, each sample was rated by only one worker. 

\myparagraph{Software} We used Python 3.11.3, Pytorch 2.0.1 and HuggingFace Transformers 4.29.2.

\myparagraph{Hardware} All experiments were run on NIVIDIA A100 GPU's with 80GB memory.

\myparagraph{Time to Run Expereiments} Experimentation time for the AMT datasets ranged from $8 - 72$hours, while time for the BLOG experimentation ranged from $48 - 168$ hours.

\section{Constrained Diverse Beam Search Algorithm and Extra Information}
Algorithm 1 is the algorithms used in Constrained Diverse Beam Search algorithm (\cdbs) proposed in our paper. It combines Diverse and Lexically Constrained Beam Search to provide a diverse candidate pool of generations that are also constrained by provided keywords. 

\begin{algorithm}\label{appx:algorithms}
\caption{Constrained-Diverse-Beam-Search (\cdbs)}
\label{alg:DCBS}
   \begin{algorithmic}
       \Require max length $n$, number of beams $k$, input ids $I$, model $M$, constraints
       \State DPP = Diverse-Preprocessing (\cref{alg:DPP})
       \State CBS = Constrained Beam Search
       \State \textit{Initialize:} $\text{beams}_0 = I$
       \For {$t = 0,...,n-1$}
            \State $\text{logits}_t = M(\text{beams}_t)$
            \State $\text{processed\_logits}_t = $ DPP($k$,  logits)
            \State $\text{beams}_{t+1} =$ CBS($\text{processed\_logits}_t$, constraints)
        \EndFor
        \State return $\text{beams}_n$
    \end{algorithmic}
\label{alg:svrg}
\end{algorithm}

\begin{algorithm}%
\label{alg:DDS}
\caption{Diverse-Preprocessing (DPP)}
\label{alg:DPP}
   \begin{algorithmic}[1]
   \Require number of beams $k$, logit matrix (\# beams $\times$ vocab size) $L$, diversity penalization term $\lambda$
   \State bincount() = vector of frequency counts of vector
   \State max() = maximum argument in vector along a specific dimension (dim)
        \State current\_tokens = []
        \For {$i = 1,...,k$}
            \If {$i=1$}
                \State processed\_logits = $L[i,:]$
            \Else
                \State previous\_token\_freq = \\ bincount(current\_tokens)
                \State processed\_logits$[i,:]$ $= L[i,:] - \lambda$ previous\_token\_freq
            \EndIf
            \If {$i < k$}
                \State current\_tokens = \\  max(processed\_logits$[0:i,:]$, dim = 1)\
            \EndIf
        \EndFor
    \State \textbf{return} processed\_logits
\end{algorithmic}
\end{algorithm}

\myparagraph{Diverse Beam Search} Traditional beam search searches for an output sequence that maximizes the conditional probability given the input. However, beam search tends to produce similar or redundant output sequences within a beam, resulting in a lack of diversity.
Diverse Beam Search (DBS) \cite{vijayakumar2018diverse} is a variation of beam search, that encourages the selection of diverse sequences that are dissimilar to each other within a beam. DBS achieves this by adding a diversity penalty term to the beam search objective function, which penalizes the selection of sequences that are too similar to the ones already in the beam. 
Its objective function can be represented as:
\begin{align*}
    \argmax_{w \in W} P_w(y\vert x) + \lambda D(y,Y)
\end{align*}
where $x$ is the sequence of previous tokens, $D(y, Y)$ is a diversity term measuring the dissimilarity between the output sequence $y$ and the set of previously selected sequences $Y$ within the beam, $\lambda$ is a hyperparameter controlling the weight of the diversity term, and $w \in W$ is the parameter vector.

The diversity penalty term can take many forms, but one common approach is to use a measure of dissimilarity such as Hamming distance or cosine similarity. By promoting diversity, Diverse Beam Search can generate more varied outputs.

\myparagraph{Constrained Beam Search} Constrained Beam Search (CBS) \cite{post2018fast} is another variant of beam search used to impose constraints on the output sequences. 
CBS achieves this by modifying the beam search objective function to penalize candidates that violate the constraints. 
The objective function for constrained beam search can be represented as:
\begin{align*}
    \argmax_{w \in W} P_w(y\vert x) + \lambda C(y)
\end{align*}
where $C(y)$ is a constraint function quantifying the degree to which the output sequence $y$ satisfies linguistic or stylistic constraints, 
and $\lambda$ is a hyperparameter controlling the weight of the constraint function.
We specifically use \textit{Lexically Constrained Beam Search} where constraints are specific words or phrases that must be included in the generated text. Concretely, while choosing candidates to fill in the beam, CBS first sorts candidates into "banks" based on number of satisfied constraints, and then selects the top $k$ candidates by iteratively visiting each bank and choosing those with the highest likelihood until reaching $k$ candidates. In terms of authorship obfuscation, we find that CBS effectively generates text closely resembling the original content by enforcing keyword inclusion, but fails to produce a variety of generations with diverse writing styles.

 \end{document}